\documentclass{article}
\usepackage{ijcai25}

\usepackage{times}
\usepackage{soul}
\usepackage{url}

\usepackage[colorlinks=true, linkcolor=blue, citecolor=blue, urlcolor=blue]{hyperref} 

\usepackage[utf8]{inputenc}
\usepackage[small]{caption}
\usepackage{graphicx}
\usepackage{amsmath}
\usepackage{amsthm}
\usepackage{amssymb}
\usepackage{booktabs}
\usepackage{algorithm}
\usepackage{algorithmic}
\usepackage[switch]{lineno}

\usepackage{bm}
\usepackage{dsfont}
\usepackage{mathtools}
\usepackage[subrefformat=parens]{subcaption}
\usepackage{array}
\usepackage{arydshln}
\usepackage{multirow}
\usepackage{csquotes}
\usepackage[enable]{easy-todo}

\usepackage{microtype}

\newcommand{\eg}{e.g.}
\newcommand{\ie}{i.e.}

\makeatletter
\newcommand{\ALC@uniqueautorefname}{Algorithm}
\makeatother

\title{Constrained Preferential Bayesian Optimization\\
and Its Application in Banner Ad Design}

\author{
Koki Iwai$^1$\and
Yusuke Kumagae$^1$\and
Yuki Koyama$^2$\and
Masahiro Hamasaki$^2$\And
Masataka Goto$^2$\\
\affiliations
$^1$Hakuhodo DY Holdings Inc.\\
$^2$National Institute of Advanced Industrial Science and Technology (AIST)\\
\emails
\{koki.iwai, yusuke.kumagae\}@hakuhodo.co.jp,
\{koyama.y, masahiro.hamasaki, m.goto\}@aist.go.jp
}

\hypersetup{
    pdftitle={Constrained Preferential Bayesian Optimization and Its Application in Banner Ad Design},
    pdfauthor={Koki Iwai, Yusuke Kumagae, Yuki Koyama, Masahiro Hamasaki, and Masataka Goto}
}

\newcommand{\figgardnersyntheticfunction}{
  \begin{figure}[t]
    \centering
    \includegraphics[keepaspectratio, width=\linewidth]{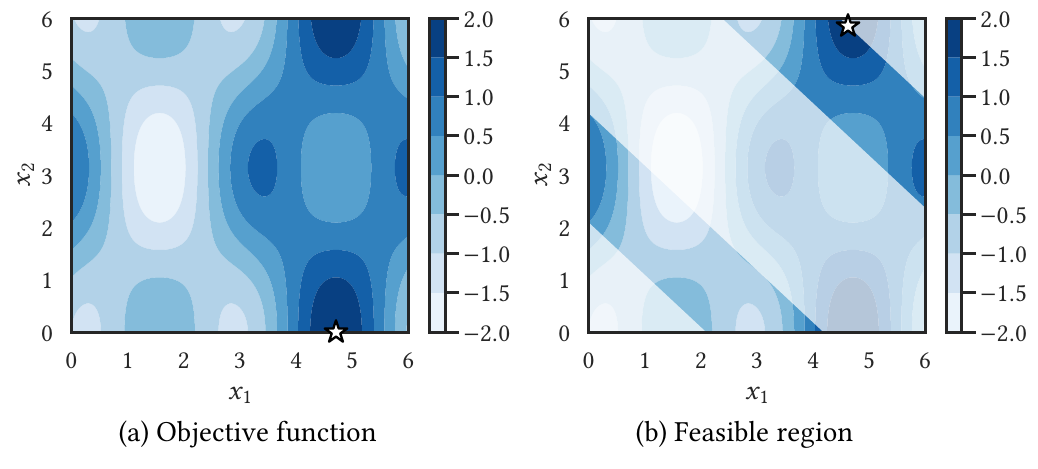}
    \caption{
      Two-dimensional test function used for evaluation.
      (a) Objective function. The star represents the maximum.
      (b) Constraint function overlaid onto the objective function. The lightly shaded white areas indicate infeasible regions, and the star represents the optimal solution that satisfies the constraint.
    }
    \label{fig:gardner_synthetic_function}
  \end{figure}
}

\newcommand{\figsyntheticresults}{
  \begin{figure*}[t]
    \centering
    \includegraphics[keepaspectratio, width=\linewidth]{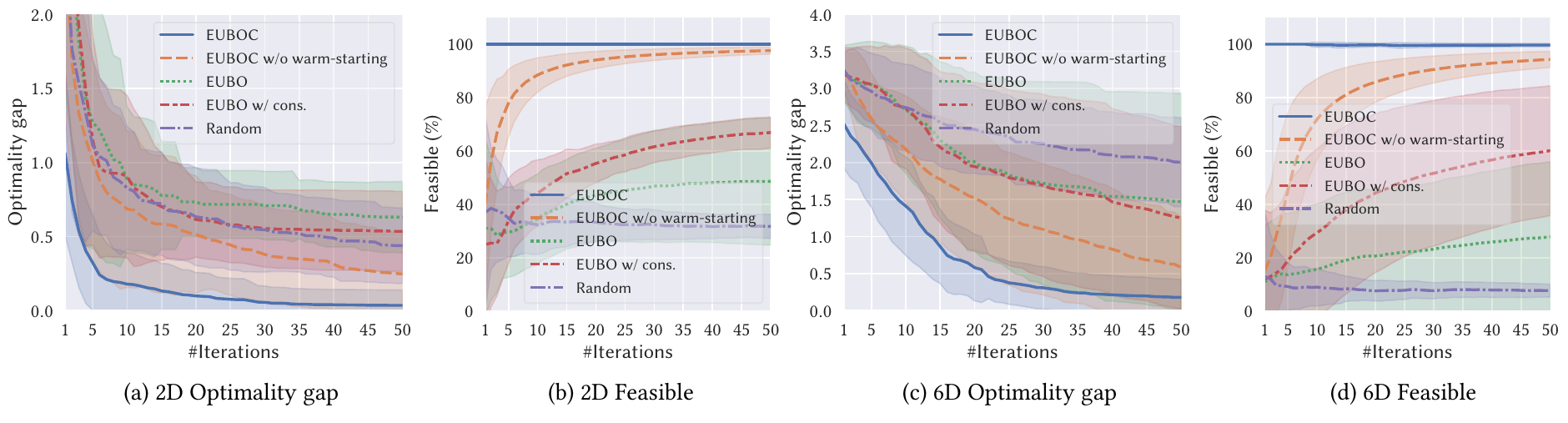}
    \caption{
      Result of the test function setting.
      The \emph{Optimality gap} for (a) 2D and (c) 6D test functions, where the horizontal axis represents the iteration steps and the vertical axis represents the \emph{Optimality gap} (lower is better).
      The \emph{Feasible} for (b) 2D and (d) 6D test functions, where the vertical axis represents the \emph{Feasible} (higher is better), the lines denote the mean, and the lightly shaded areas denote the standard deviation.
    }
    \label{fig:synthetic_results}
  \end{figure*}
}

\newcommand{\figsystemresults}{
  \begin{figure}[t]
    \centering
    \includegraphics[keepaspectratio, width=\linewidth]{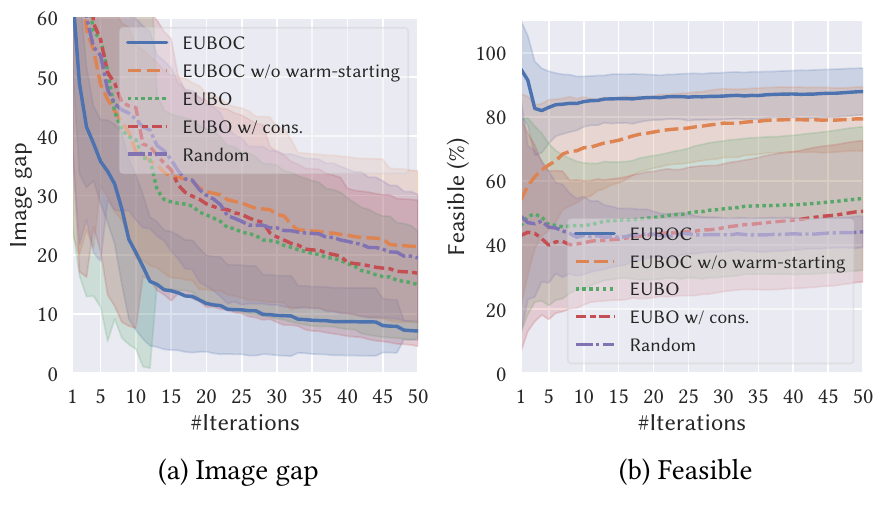}
    \caption{
      Result of the banner ad design application setting. (a) \emph{Image gap} (lower is better) and (b) \emph{Feasible} (higher is better).
    }
    \label{fig:system_results}
  \end{figure}
}

\newcommand{\figureframework}{
    \begin{figure}[t]
      \includegraphics[width=\linewidth]{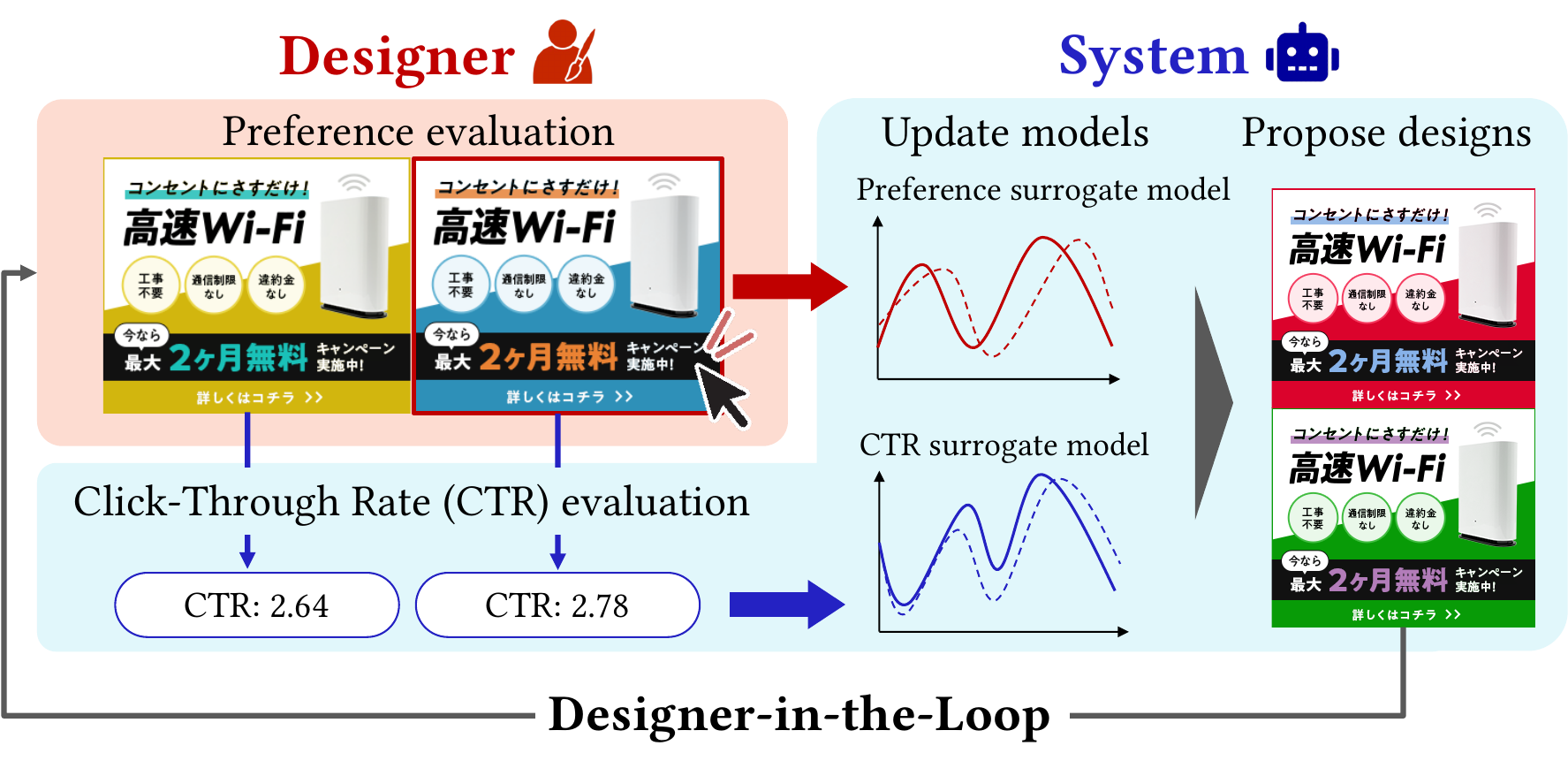}
      \caption{
        Concept of our designer-in-the-loop banner ad design framework.
        The designer provides feedback to the system about preferences, and the system predicts the CTRs.
        The system then updates the surrogate models and proposes new design candidates using our CPBO technique.
      }
      \label{fig:framework}
    \end{figure}
}

\newcommand{\figuserinterface}{
  \begin{figure}[t]
    \centering
    \includegraphics[keepaspectratio, width=\linewidth]{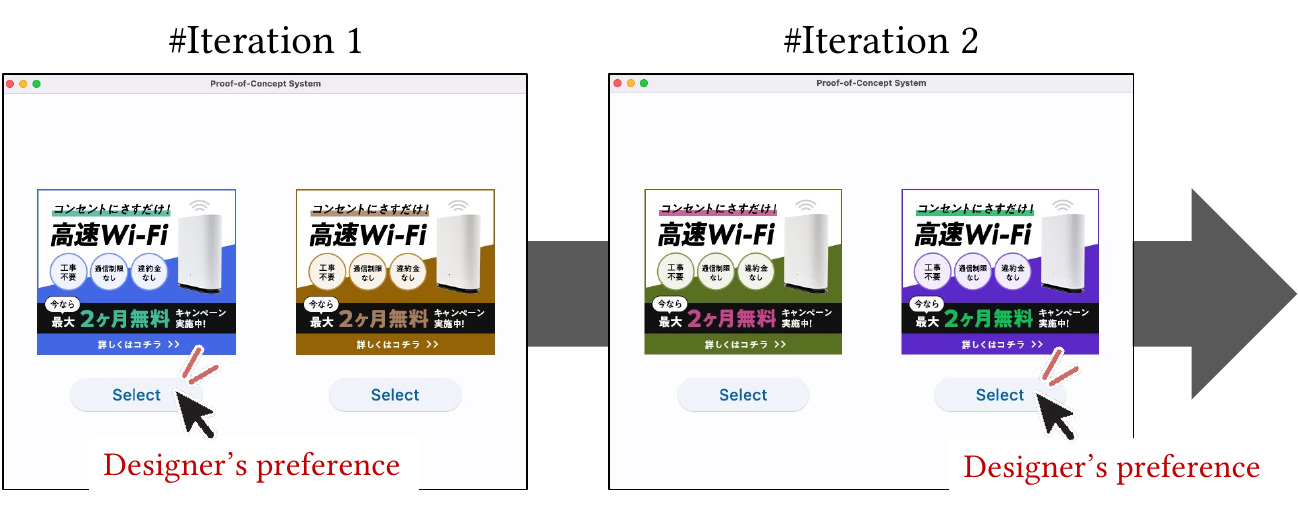}
    \caption{
      User interface of our banner ad design system in color editing mode.
      This system provides pairs of design candidates for each iteration.
    }
    \label{fig:user_interface}
  \end{figure}
}

\newcommand{\figquestionarrie}{
  \begin{figure}[t]
    \centering
    \includegraphics[keepaspectratio, width=\linewidth]
    {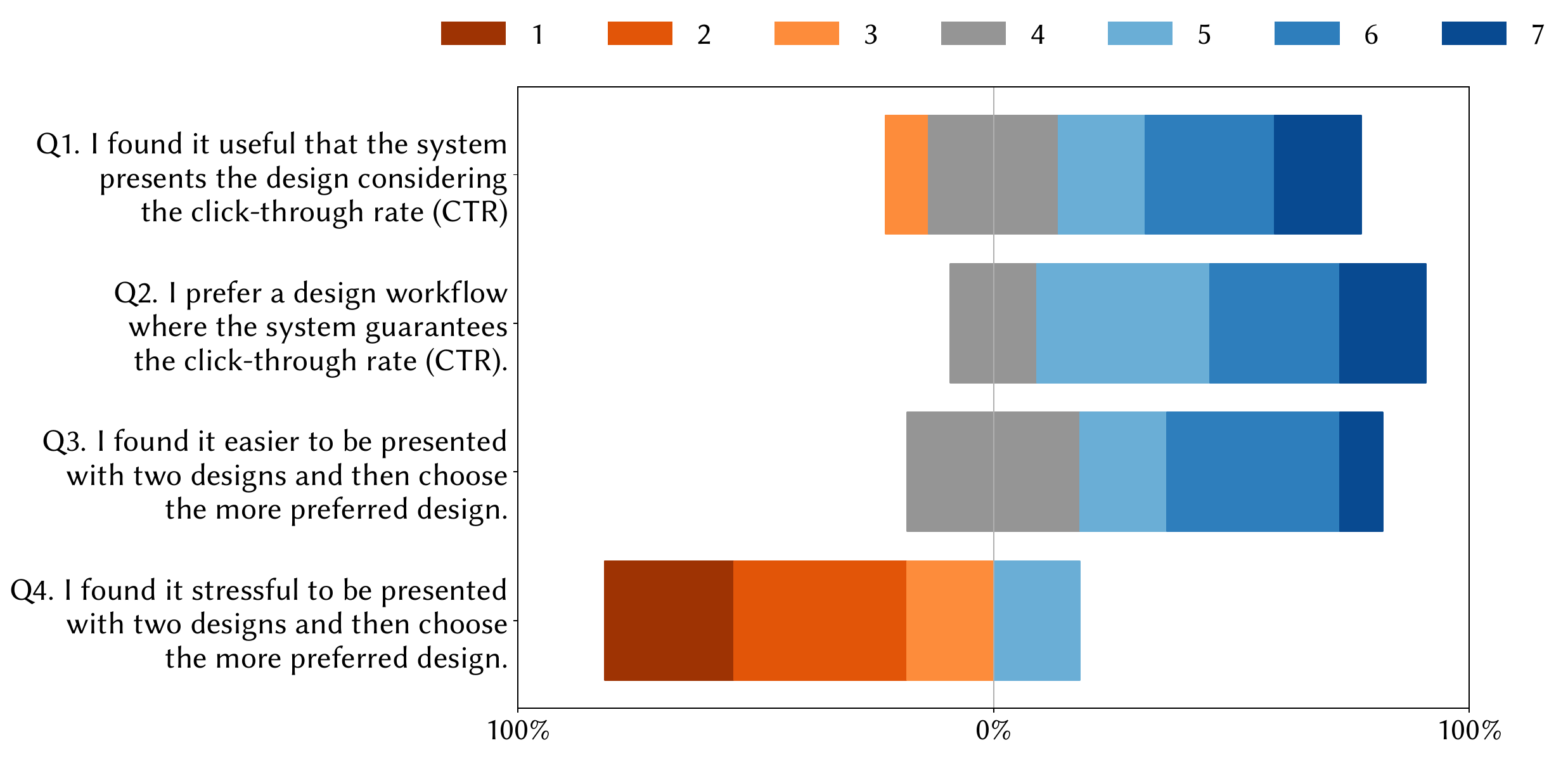}
    \caption{
      Questionnaire results, showing the distribution of answers for each questionnaire item.
      Q1 to Q3 are better to the right, and Q4 is better to the left.
    }
    \label{fig:questionarrie}
  \end{figure}
}

\newcommand{\figwarmstartresults}{
  \begin{figure*}[t]
    \centering
    \includegraphics[keepaspectratio, width=\linewidth]{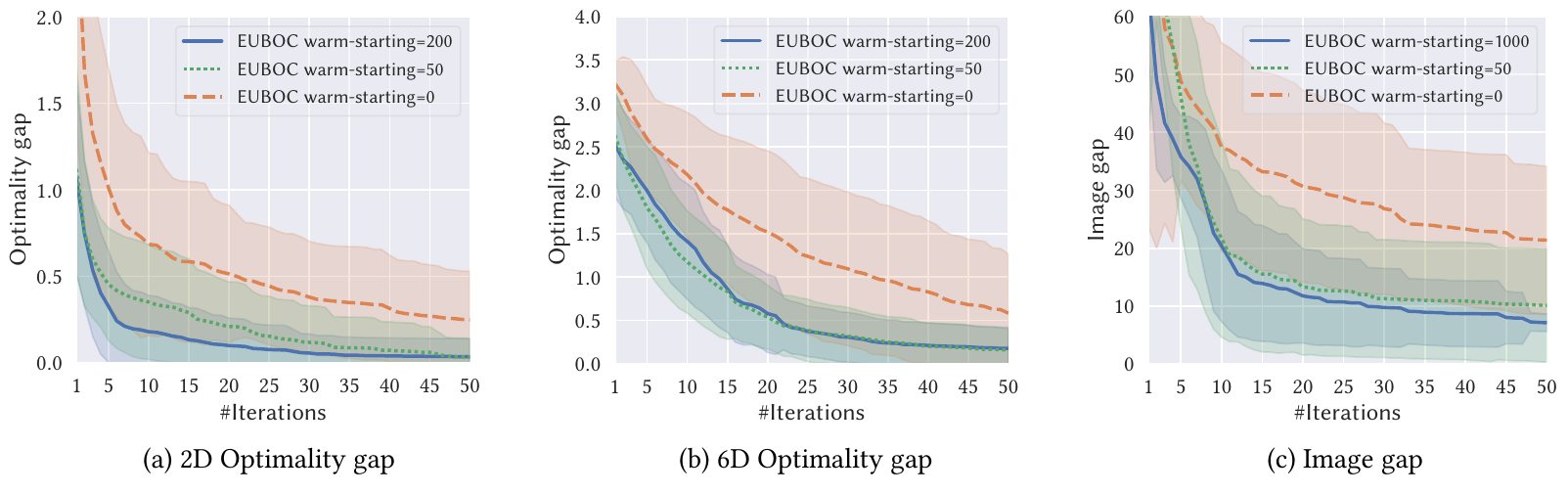}
    \caption{
      Performance of EUBOC with 50 warm-start points. (a) \emph{Optimality gap} in the 2D test function setting, (b) \emph{Optimality gap} in the 6D test function setting and (c) \emph{Image gap} in the application setting.
    }
    \label{fig:warm-start_results}
  \end{figure*}
}

\newcommand{\figreferenceimage}{
    \begin{figure}[t]
        \centering
        \includegraphics[keepaspectratio, width=0.4\linewidth]{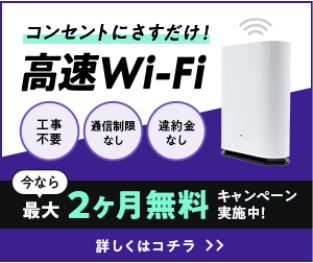}
        \caption{Reference image used in technical evaluation.}
        \label{fig:banner_reference}
    \end{figure}
}

\newcommand{\figtechadresults}{
    \begin{figure*}[htp]
        \centering
        \includegraphics[keepaspectratio, width=0.95\linewidth]{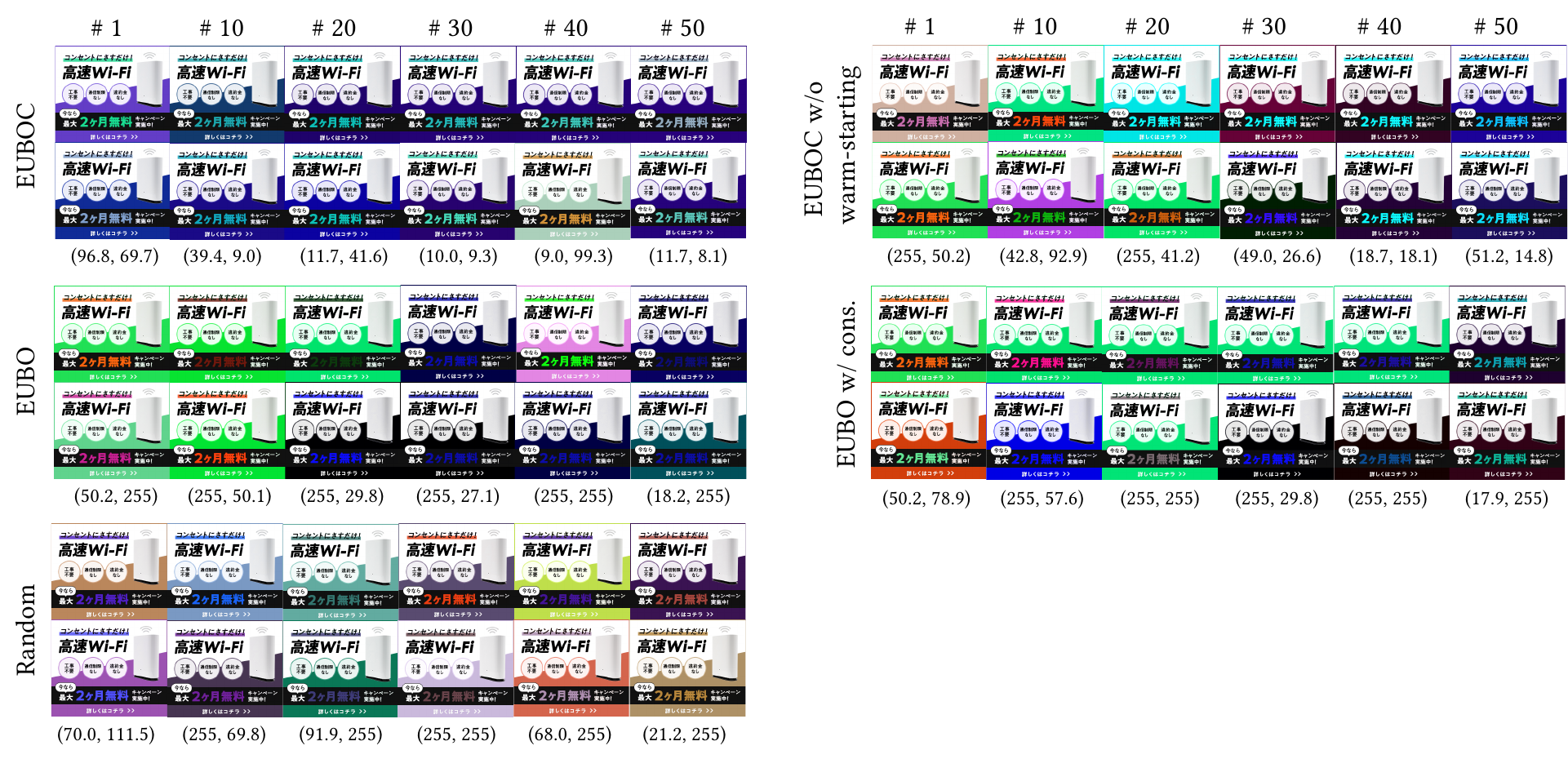}
        \caption{Ad designs in banner ad design application setting results. Each method displays two ad images per iteration, stacked vertically as the compared pair. The numbers below the images represent the \emph{Image gap}: the left number for the top image and the right number for the bottom image (lower is better). An image gap of 255 indicates that the corresponding image does not satisfy the constraint.}
        \label{fig:ad_results_tech_eval}
    \end{figure*}
}

\newcommand{\figuserinterfaceadd}{
  \begin{figure*}[t]
    \centering
    \includegraphics[keepaspectratio, width=\linewidth]{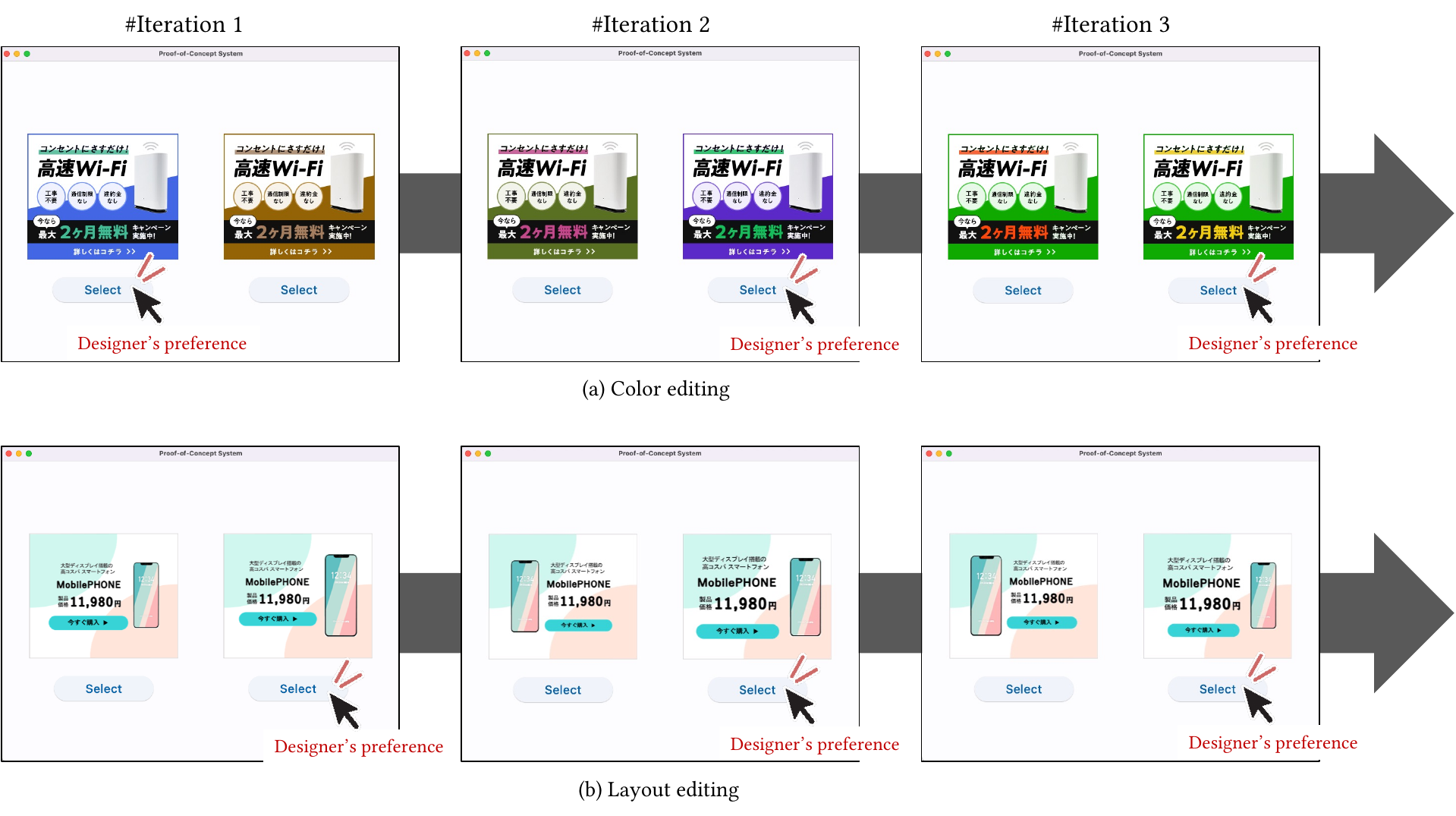}
    \caption{
      User interface of our banner ad design system in (a) color editing mode and (b) layout editing mode.
      This system provides pairs of design candidates for each iteration.
    }
    \label{fig:user_interface_add}
  \end{figure*}
}

\newcommand{\figquestionnairers}{
  \begin{figure}[t]
    \centering
    \includegraphics[keepaspectratio, width=\linewidth]{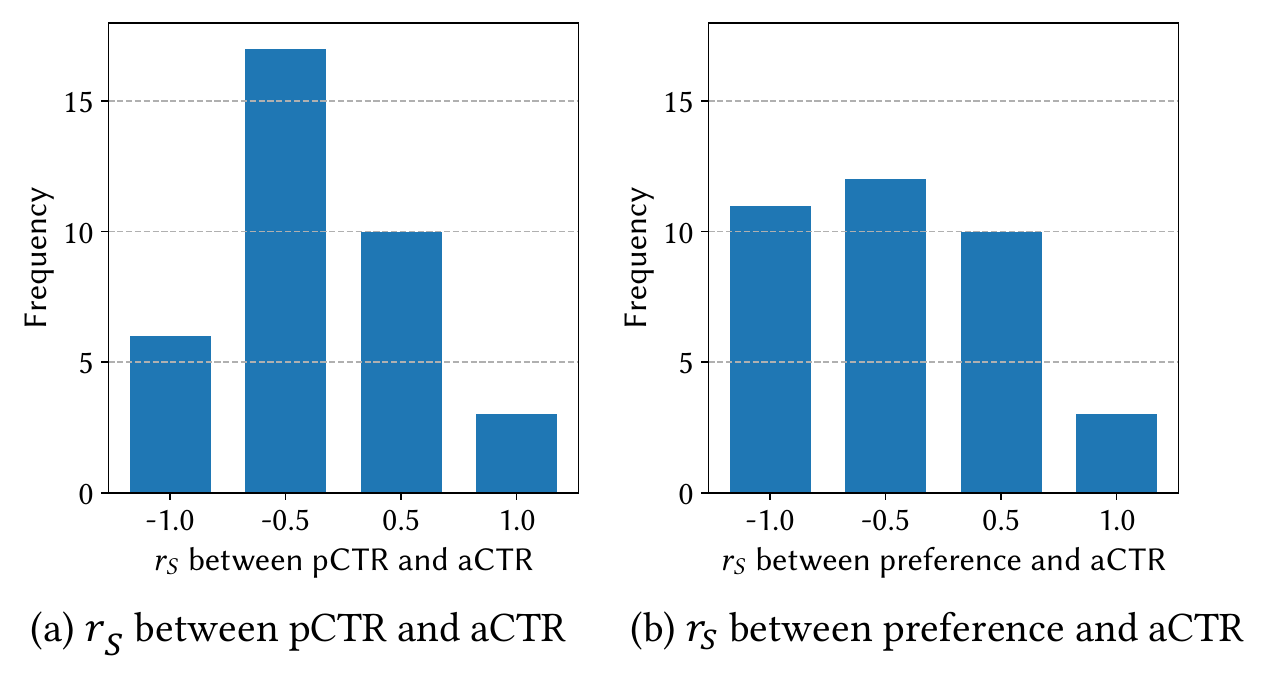}
    \caption{
      Histogram of Spearman's rank correlation coefficients (a) between predicted CTR (pCTR) and actual CTR (aCTR) and (b) between preference and aCTR.
    }
    \label{fig:questionnaire_rs}
  \end{figure}
}

\newcommand{\figbanner}{
  \begin{figure}[t]
    \centering
    \includegraphics[keepaspectratio, width=\linewidth]{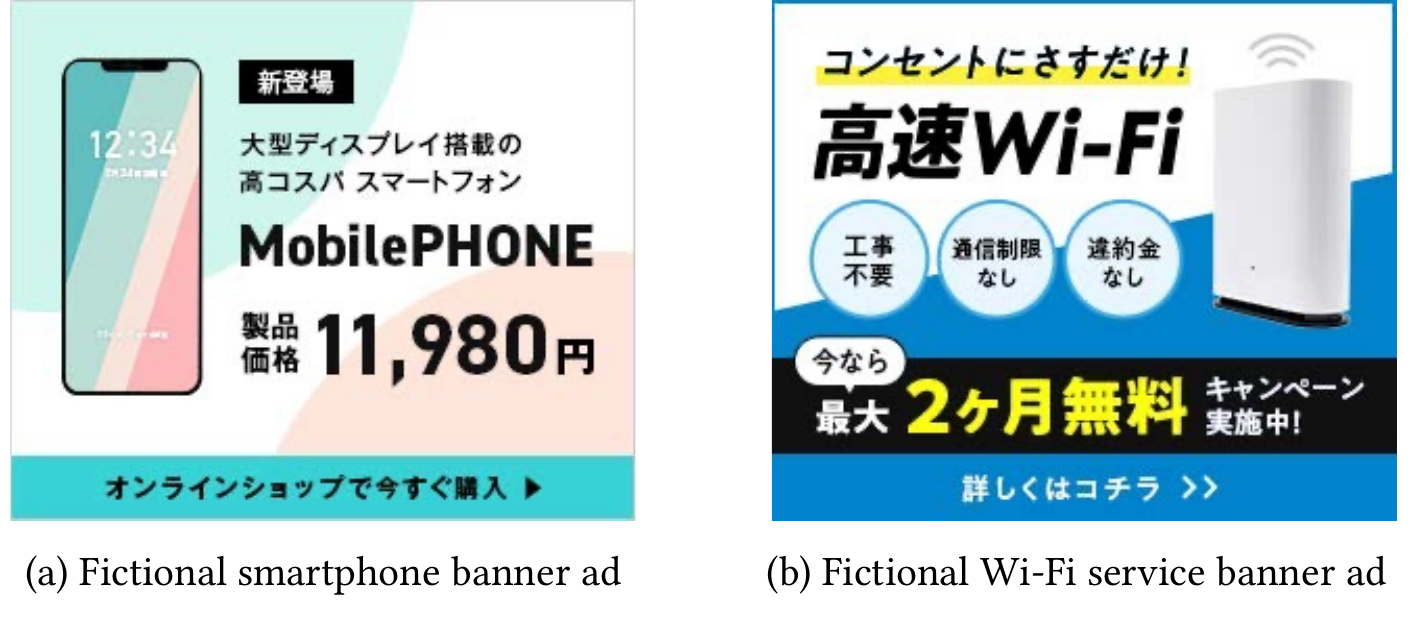}
    \caption{
      Banner ad images used in the user study.
      These were created specifically for this study by a professional ad designer.
      (a) Fictional smartphone banner ad.
      (b) Fictional Wi-Fi service banner ad.
    }
    \label{fig:banner}
  \end{figure}
}

\newcommand{\figuseradresults}{
    \begin{figure*}[htp]
        \centering
        \includegraphics[keepaspectratio, width=0.95\linewidth]{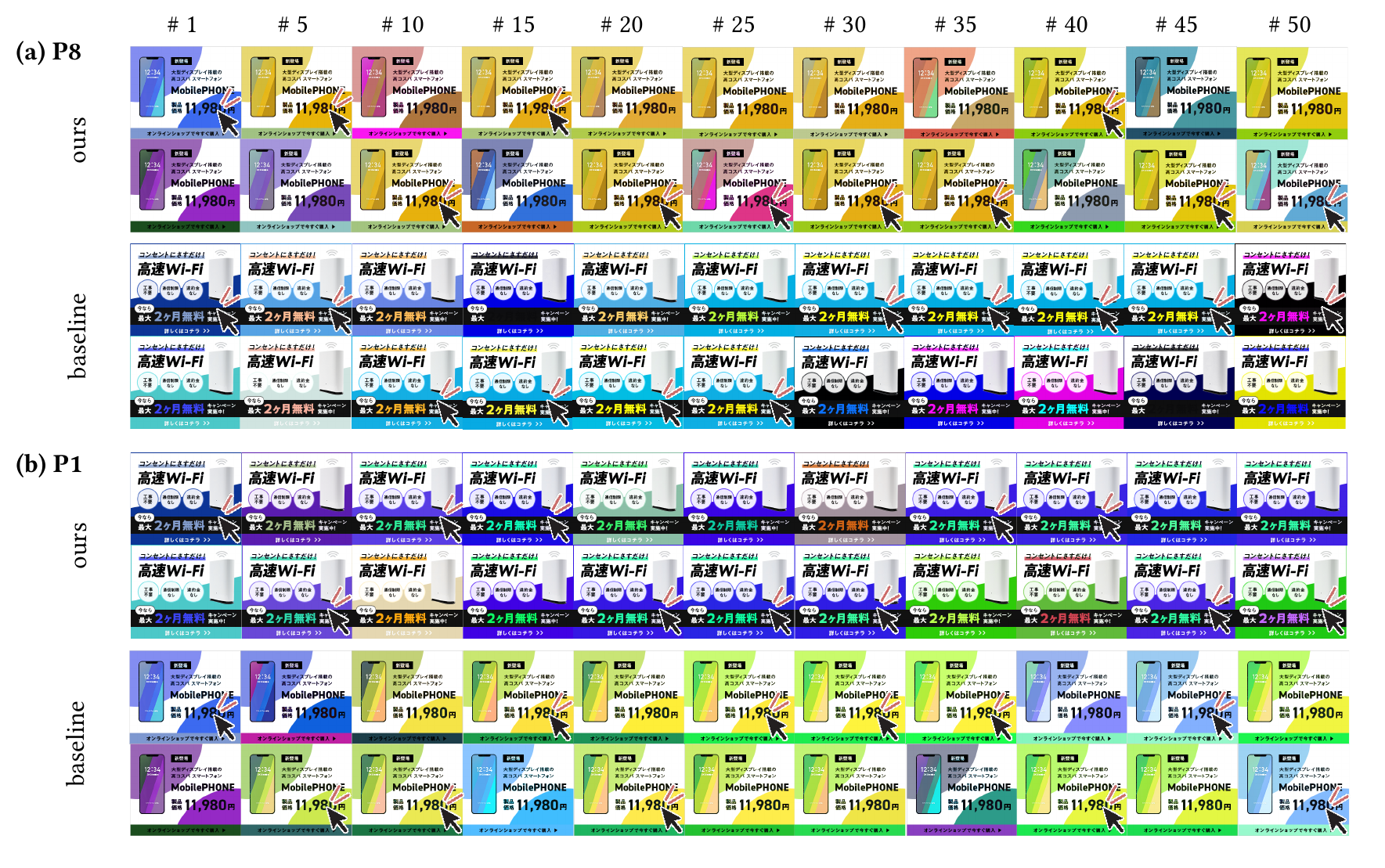}
        \caption{Candidate ad designs and selection results for our user study conducted by (a) P8 and (b) P1.}
        \label{fig:ad_results_user_eval}
    \end{figure*}
}

\newcommand{\figfollowupbanner}{
    \begin{figure}[t]
        \centering
        \includegraphics[keepaspectratio, width=\linewidth]{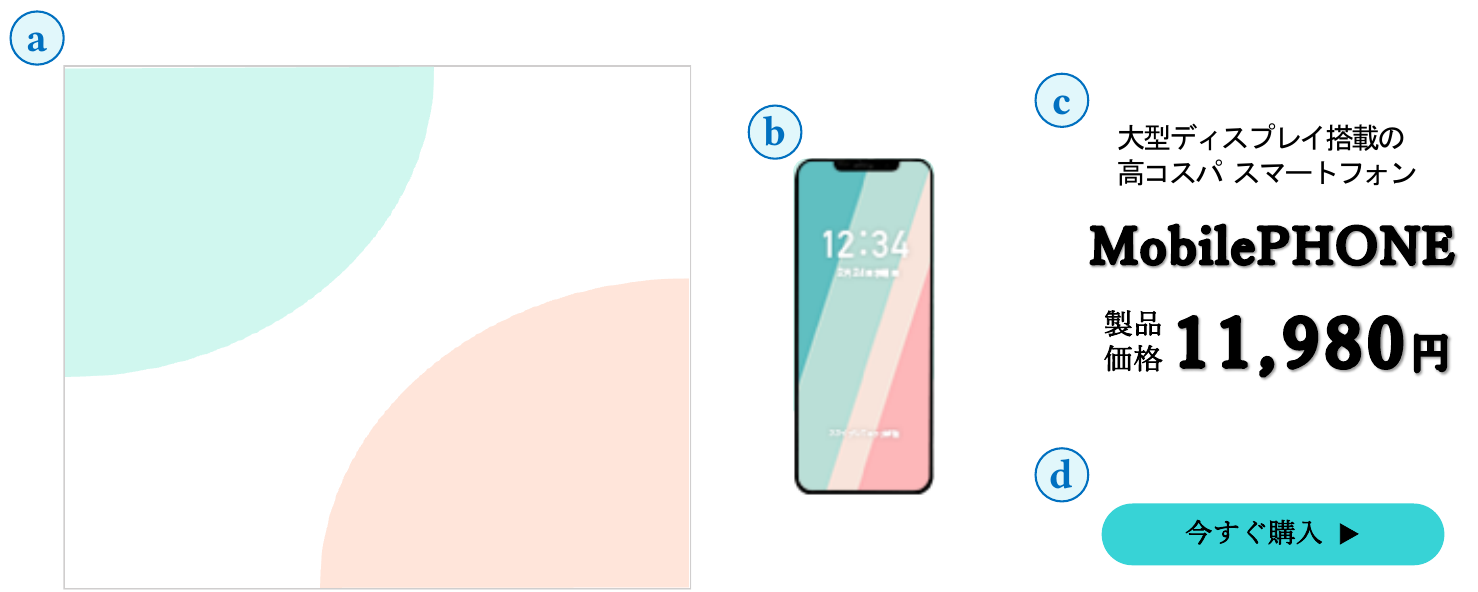}
        \caption{Elements of a banner ad used in our follow-up user study. The banner ad is composed of (a) a background image, (b) a product image, (c) an advertising text, and (d) a call-to-action button}
        \label{fig:ad_follow_up}
    \end{figure}
}

\newcommand{\figfollowupresults}{
    \begin{figure*}[htp]
        \centering
        \includegraphics[keepaspectratio, width=\linewidth]{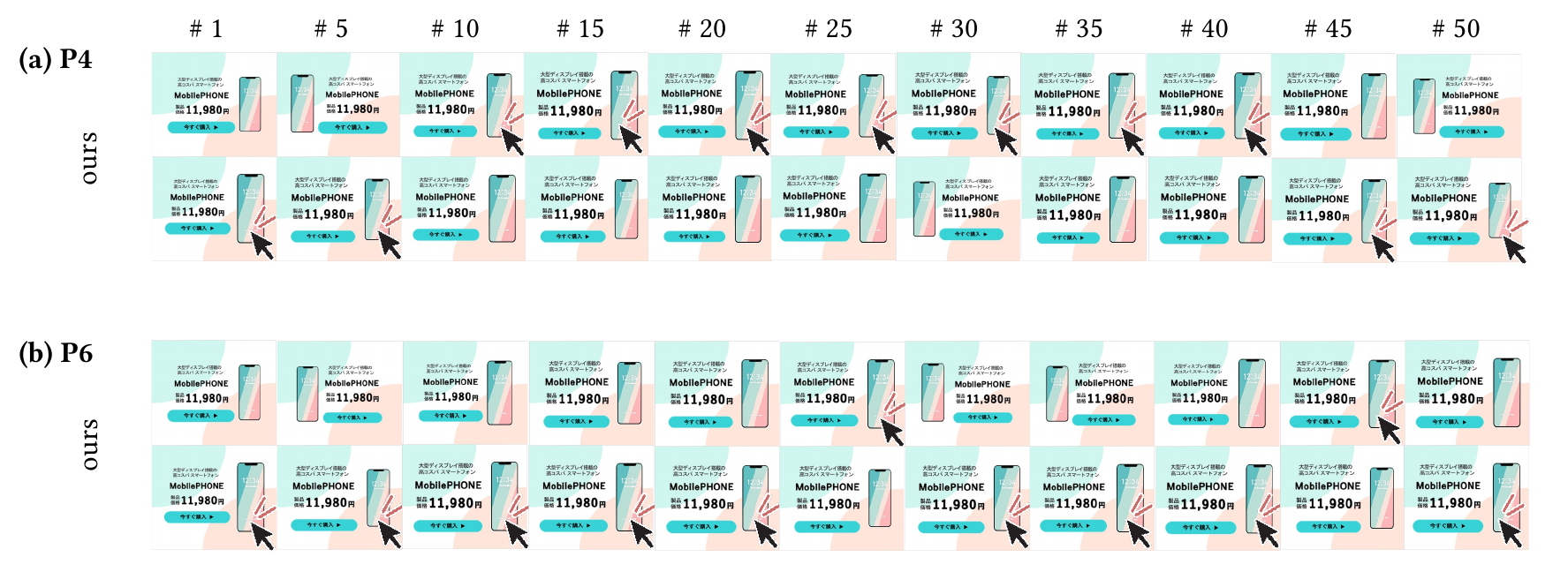}
        \caption{Candidate ad designs and selection results for our follow-up user study conducted by (a) P4 and (b) P6.}
        \label{fig:ad_results_follow_up}
    \end{figure*}
}

\begin{document}

\maketitle

\renewcommand{\thefootnote}{}\footnotemark
\footnotetext{This paper has been accepted at IJCAI 2025. This version includes supplementary material in the appendix.}
\renewcommand{\thefootnote}{\arabic{footnote}}

\begin{abstract}

Preferential Bayesian optimization (PBO) is a variant of Bayesian optimization that observes relative preferences (\eg, pairwise comparisons) instead of direct objective values, making it especially suitable for human-in-the-loop scenarios.
However, real-world optimization tasks often involve inequality constraints, which existing PBO methods have not yet addressed.
To fill this gap, we propose \emph{constrained preferential Bayesian optimization} (CPBO), an extension of PBO that incorporates inequality constraints for the first time.
Specifically, we present a novel acquisition function for this purpose.
Our technical evaluation shows that our CPBO method successfully identifies optimal solutions by focusing on exploring feasible regions.
As a practical application, we also present a designer-in-the-loop system for banner ad design using CPBO, where the objective is the designer's subjective preference, and the constraint ensures a target predicted click-through rate.
We conducted a user study with professional ad designers, demonstrating the potential benefits of our approach in guiding creative design under real-world constraints.

\end{abstract}


\section{Introduction}
\label{sec:intro}

Preferential Bayesian optimization (PBO) \cite{Brochu2007-xw,Gonzalez2017-hu,Koyama2022-by} is a variant of Bayesian optimization (BO) \cite{Shahriari2016-si} that observes relative preferences (\eg, pairwise comparisons) instead of direct objective values.
Since relative evaluations are generally considered easier, faster, and more accurate than absolute evaluations for human subjective assessments \cite{Brochu2010-qi,Yoshida2024-sq}, PBO is particularly well-suited for problems in which human preference serves as the objective function to be maximized.
It has been effectively employed to implement human-in-the-loop optimization systems for visual design \cite{Brochu2007-xw,Brochu2010-qi,Koyama2017-yz,Koyama2020-pm,Yamamoto2022-vi}.

However, existing PBO methods have not been adapted to handle constrained optimization problems.  
Many real-world optimization tasks (\eg, product and architectural design, drug discovery, and recommender systems) involve maximizing human preferences under additional constraints.  
For example, in product and architectural design, users may express preferences over usability, while the design must also satisfy constraints such as durability or thermal performance, often evaluated through physical simulations.  
These scenarios require human-in-the-loop optimization under costly or implicit constraints, highlighting a technical gap in extending PBO to such applications.

To address this gap, we propose a new method called \emph{constrained preferential Bayesian optimization} (CPBO), which incorporates inequality constraints into PBO.
As the core of CPBO, we introduce a novel acquisition function, \emph{expected utility of the best option with constraints} (EUBOC), which extends an existing PBO acquisition function \cite{Lin2022-et}.
We evaluate the proposed method through simulation experiments, highlighting its ability to find optimal solutions while effectively accounting for constraints.

As a practical application of CPBO, we propose a novel \emph{designer-in-the-loop} framework for banner ad design, where the predicted \emph{click-through rate} (CTR) serves as a constraint.
Banner ads are used to promote products or services online, and CTR, the fraction of clicks relative to impressions, is a key metric in the advertising industry \cite{McMahan2013-jw,Chen2016-kq,Richardson2007-qh,Zhou2018-hy}.
However, focusing solely on maximizing CTR can lead to designs that, while effective in capturing clicks, may be visually unappealing or even annoying, negatively affecting brand perception \cite{Zeng2020-qm,Zeng2021-em}.
Consequently, effective banner ad design requires maximizing visual appeal while maintaining a reasonably high CTR.
This task is more complex than typical visual design tasks that focus solely on subjective preferences, such as those addressed in previous work \cite{Brochu2010-qi,Koyama2020-pm,Yamamoto2022-vi}.
To evaluate its effectiveness, we conducted a user study with professional ad designers, revealing that they positively received the concept of our framework and appreciated its potential to reduce design workload.

Our contributions are summarized as follows.
\begin{itemize}
    \item We propose CPBO, a novel extension of PBO that handles inequality-constrained optimization problems.
    \item To implement CPBO, we propose EUBOC, a new acquisition function for CPBO. We evaluate its performance through simulation experiments.
    \item As a practical CPBO application, we propose a novel designer-in-the-loop framework for banner ad design that integrates CTR considerations. A user study with professional ad designers validates its real-world viability.
\end{itemize}


\section{Preliminaries}
\label{sec:prelinimaries}

Before describing our proposed CPBO technique, we describe its foundation: PBO and constrained Bayesian optimization (CBO).
Especially, we detail their acquisition functions: \emph{expected utility of the best option} (EUBO)~\cite{Lin2022-et} for PBO, and \emph{expected improvement with constraints} (EIC)~\cite{Gardner2014-tv} for CBO.

Throughout this paper, we assume that the input variables consist of $N$ continuous parameters, each normalized to the range of $[0,1]$ (without loss of generality).
We denote these parameters by $x_i \in [0,1]$ ($i=1,\ldots, N$), and define the vector $\bm{x} = [x_1, \ldots, x_N] \in \mathcal{X}$, where $\mathcal{X}=[0,1]^N$.
We further denote the objective function by $f: \mathcal{X} \to \mathbb{R}$ and the constraint function by $c: \mathcal{X} \to \mathbb{R}$.

\subsection{Preferential Bayesian Optimization}
\label{sec:preliminaries:pbo}

PBO is a variant of BO that observes relative preferences instead of direct objective function values.
As with standard BO, the goal of PBO is to identify the global optimum of a black-box function:
\begin{align}
  \bm{x}^* = \mathop{\operatorname{arg\ max}}_{\bm{x} \in \mathcal{X}} f(\bm{x}).
\end{align}
However, PBO achieves this by iteratively collecting preference data instead of observing the function value $f(\bm{x})$ directly.
In this work, we assume that each observation yields a forced choice between two candidates, denoted as $d = \bm{x}^{(i)} \succ \bm{x}^{(j)}$ (\ie, the candidate $\bm{x}^{(i)}$ is preferred over the candidate $\bm{x}^{(j)}$).

PBO is often used in human-in-the-loop settings~\cite{Koyama2022-by}, where the black-box objective function represents a human preference, and the goal is to find the most preferred option through iterative human evaluations.

In PBO, a probabilistic model serves as a surrogate for the objective function $f$, and this model is continuously updated based on the observed preference data through Bayesian inference.
Gaussian processes (GPs)~\cite{Rasmussen2005-nn} are often used as surrogates owing to their flexibility;
we adopt GPs for this purpose as well.

The likelihood of a preference data $d = \bm{x}^{(i)} \succ \bm{x}^{(j)}$ is often modeled using the Thurstone-Mosteller model~\cite{Chu2005-wm}:
\begin{equation}
P(d\,|\,f)=\Phi\left(\frac{f(\bm{x}^{(i)})-f(\bm{x}^{(j)})}{\sqrt{2}\sigma}\right),
\end{equation}
where $\Phi(\cdot)$ is the cumulative distribution function of the standard normal distribution, and $\sigma$ is the standard deviation of the Gaussian noise.
The likelihood of multiple preference data, $\mathcal{D}=\{d_1,d_2,\ldots\}$ is given by $P(\mathcal{D}|f)=\Pi_{i}P(d_i|f)$.
Using this probability, the surrogate model can be updated; refer to \cite{Chu2005-wm} for details.

At each iteration of PBO, the next candidate point to be compared is selected by maximizing an acquisition function derived from the updated surrogate model.
The acquisition function measures a candidate's effectiveness, and its design is critical to the overall performance of PBO.

EUBO~\cite{Lin2022-et} is one of the state-of-the-art acquisition functions for PBO.
It can be expressed in closed form for two search points $\bm{x}^{(i)}$ and $\bm{x}^{(j)}$ as
\begin{align}
\text{EUBO}(\bm{x}^{(i)},\,\bm{x}^{(j)}) &= \mathbb{E}[\textrm{max}\{f(\bm{x}^{(i)}),\,f(\bm{x}^{(j)})\}] \nonumber \\
&=\Delta\Phi\left(\frac{\Delta}{\sigma}\right)+\sigma\phi\left(\frac{\Delta}{\sigma}\right)+\mu^f(\bm{x}^{(j)}),
\end{align}
where $\phi(\cdot)$ is the probability density function of the standard normal distribution, $\Delta=\mathbb{E}[f(\bm{x}^{(i)})-f(\bm{x}^{(j)})]$, $\sigma^2=\text{Var}[f(\bm{x}^{(i)})-f(\bm{x}^{(j)})]$, and $\mu^f(\bm{x}^{(j)})=\mathbb{E}[f(\bm{x}^{(j)})]$.
We adopt EUBO as the basis of our proposed technique because of its efficiency and computational simplicity.

\subsection{Constrained Bayesian Optimization}

CBO is a variant of BO designed for optimization problems with an additional inequality constraint defined as
\begin{align}
c(\bm{x}) \geq \lambda,
\end{align}
where $c$ is a black-box constraint function and $\lambda$ denotes a threshold for the constraint.
In addition to modeling the objective function, CBO typically constructs a GP surrogate for the constraint function;
we adopt this approach in our work.

EIC~\cite{Gardner2014-tv} is an acquisition function for CBO.
The key idea is to multiply the \emph{expected improvement} (EI), a popular acquisition function for BO, by the probability of satisfying the constraint;
that is,
\begin{equation}
\text{EIC}(\bm{x}) = P(c(\bm{x})\geq\lambda)\,\text{EI}(\bm{x}).
\label{EIC_formulation}
\end{equation}
We adopt EIC as the constraint-handling method in our proposed technique because of its simplicity and compatibility with EUBO.
Although many constraint-handling methods exist~\cite{Hernandez-Lobato2015-dm,Amini2025-vs}, most are \emph{incompatible} with EUBO, such as those requiring direct objective values or information-theoretic assumptions.
Given these considerations, we find EIC to be a suitable choice.


\section{Constrained Preferential Bayesian Optimization}
\label{sec:technique}

\subsection{Problem Formulation}

We aim to solve the following constrained optimization problem:
\begin{equation}
\bm{x}^* = \mathop{\operatorname{arg\ max}}_{\bm{x} \in \mathcal{X}} f(\bm{x}) \quad \text{s.t.} \quad c(\bm{x}) \geq \lambda,
\label{eq:problem_formulation}
\end{equation}
where $\lambda$ denotes a threshold for the constraint.
Unlike standard BO, the objective function value $f(\bm{x})$ is not directly observed;
instead, we observe relative preferences among multiple candidates to infer the objective function, as in PBO methods (\autoref{sec:preliminaries:pbo}).
In contrast, we assume that the constraint function value $c(\bm{x})$ can be directly observed for any given $\bm{x}$ without requiring human feedback.
Following the original EIC assumptions~\cite{Gardner2014-tv}, we treat $c(\bm{x})$ as a black-box function that is expensive to evaluate, noise-free, and does not prevent evaluation of the objective when violated.

In practical scenarios, the objective function $f$ often represents human preferences for a design (\ie, visual appeal), while the constraint function $c$ captures design feasibility (\eg, performance requirements).
Accordingly, we estimate $f$ by iteratively asking human evaluators for relative preferences among multiple provided candidates, thereby capturing subjective preferences that would otherwise be difficult to quantify.

\subsection{Acquisition Function: EUBOC}

\begin{algorithm}[t]
    \small
    \caption{
        CPBO with pairwise comparison
    }
    \begin{algorithmic}[1]
        \STATE $\mathcal{H} \leftarrow \varnothing$
        \FOR {$n=1,2,\ldots$}
        \STATE select $\bm{x}_{n}^{(i)}, \bm{x}_{n}^{(j)}$ by optimizing EUBOC:
        $$
        \bm{x}_{n}^{(i)},\, \bm{x}_{n}^{(j)} = \mathop{\operatorname{arg\ max}}_{\bm{x}^{(i)},\ \bm{x}^{(j)} \in \mathcal{X}} {}\, \text{EUBOC}(\bm{x}^{(i)},\,\bm{x}^{(j)})
        $$
        \STATE Ask the evaluator to compare $\bm{x}_{n}^{(i)}$ and $\bm{x}_{n}^{(j)}$
        \STATE Evaluate the constraint function to obtain $c_{n}^{(i)}$ and $c_{n}^{(j)}$
        \STATE $\mathcal{H} \leftarrow \mathcal{H} \cup \{(\bm{x}_{n}^{\textrm{(selected)}} \succ \bm{x}_{n}^{\textrm{(not\,selected)}}), (c_{n}^{(i)}, c_{n}^{(j)})\}$
        \STATE Update surrogate models $f$ and $c$ based on $\mathcal{H}$
        \ENDFOR
    \end{algorithmic}
    \label{alg:pbo}
\end{algorithm}

We propose a new acquisition function for CPBO, \emph{expected utility of the best option with constraints} (EUBOC), which integrates EUBO with the idea of EIC.
Specifically, we propose multiplying EUBO by the probability that the two points are both feasible:
\begin{align}
&\text{EUBOC}(\bm{x}^{(i)},\,\bm{x}^{(j)}) \nonumber \\
&= P(c(\bm{x}^{(i)})\geq\lambda, c(\bm{x}^{(j)})\geq\lambda)\,\text{EUBO}(\bm{x}^{(i)},\,\bm{x}^{(j)}).
\label{eq:euboc}
\end{align}
Assuming a GP for the constraint function $c$, the pair $(c(\bm{x}^{(i)}),c(\bm{x}^{(j)}))$ follows a bivariate normal distribution.
Therefore, the probability of satisfying the inequality constraints is
\begin{align}
&P(c(\bm{x}^{(i)})\geq\lambda, c(\bm{x}^{(j)})\geq\lambda) \nonumber \\
&= \int_{\lambda}^{\infty}\int_{\lambda}^{\infty}P(c(\bm{x}^{(i)}),c(\bm{x}^{(j)}))dc(\bm{x}^{(i)})dc(\bm{x}^{(j)}).
\label{eq:exact}
\end{align}
Given the GP assumption, a correlation generally exists between $c(\bm{x}^{(i)})$ and $c(\bm{x}^{(j)})$.
However, deriving the cumulative distribution function values analytically for a bivariate normal distribution is challenging.
Therefore, we apply the following approximation, assuming that $c(\bm{x}^{(i)})$ and $c(\bm{x}^{(j)})$ are uncorrelated:
\begin{align}
& \autoref{eq:exact} \nonumber \\
&\approx\int_{\lambda}^{\infty}P(c(\bm{x}^{(i)}))dc(\bm{x}^{(i)})\int_{\lambda}^{\infty}P (c(\bm{x}^{(j)}))dc(\bm{x}^{(j)}) \nonumber \\
&=\left(1-\Phi\left(\frac{\lambda-\mu^c(\bm{x}^{(i)})}{\sigma^c(\bm{x}^{(i)})}\right)\right)\left(1-\Phi\left(\frac{\lambda-\mu^c(\bm{x}^{(j)})}{\sigma^c(\bm{x}^{(j)})}\right)\right),
\label{eq:cons_p}
\end{align}
where $\mu^c(\bm{x})=\mathbb{E}[c(\bm{x})]$ and $\sigma^c(\bm{x})=\sqrt{\text{Var}[c(\bm{x})}]$.

\subsection{Algorithm}

The proposed algorithm is presented in \autoref{alg:pbo}, where $\mathcal{H}$ denotes the history of the evaluation data.
At step $n$, two points are obtained by solving the maximization problem of the EUBOC acquisition function (line 3).
For the obtained two points $\bm{x}_{n}^{(i)}$ and $\bm{x}_{n}^{(j)}$, we obtain the preference data $(\bm{x}_{n}^{\textrm{(selected)}} \succ \bm{x}_{n}^{\textrm{(not\,selected)}})$ from the (human) evaluator (line 4) and also their constraint values $(c_{n}^{(i)}, c_{n}^{(j)})$ by evaluating the constraint function (line 5).
Then, we add these data to the history $\mathcal{H}$ and update the GP surrogate models $f$ and $c$ (lines 6 and 7).
This process is repeated for a certain number of iterations.

\subsection{Warm-Starting Constraint Surrogate Model Update (Optional)}
\label{sec:tech:warm}

Optionally, it is possible to pre-train the constraint surrogate model before starting the optimization iterations involving the human evaluator, given that the evaluator does not intervene in evaluating the constraint function.
This \emph{warm-start} approach facilitates starting the optimization with lower uncertainty regarding the constraints, potentially leading to improved optimization efficiency.
The efficacy of this approach is demonstrated in \autoref{sec:tech_eval}, where we pre-trained the constraint surrogate model using points randomly sampled from the search space and their constraint function values.


\section{Technical Evaluation}
\label{sec:tech_eval}

We conducted \emph{simulation} experiments to evaluate the proposed CPBO technique.
The goal was to confirm that the proposed technique can find solutions that satisfy the inequality constraint and to understand its efficiency.

\subsection{Test Function Setting}
\label{sec:synthetic_setting}

In this setting, we simulated both human responses (pairwise comparison) and constraint queries (direct observation of the constraint function values) using known synthetic test functions.

\subsubsection{Test Functions}

\figgardnersyntheticfunction

\figsyntheticresults

We employed two synthetic problems for evaluation.
The first comes from Gardner \textit{et al.}~\shortcite{Gardner2014-tv}.
It consists of 2-dimensional (2D) objective and constraint functions composed of sine and cosine functions.
\autoref{fig:gardner_synthetic_function} visualizes the objective function and feasible region.
The second comes from Letham \textit{et al.}~\shortcite{Letham2019-sm}.
It comprises 6-dimensional (6D) objective and constraint functions, representing a higher-dimensional case closer to real-world tasks.
Its objective function is based on a Hartmann 6 function~\cite{Picheny2013-kf}, while the constraint function uses the norm of $\bm{x}$ (see \autoref{sec:appendix:synth} for details).

\subsubsection{Methods to be Compared}

We compared the following methods:
\begin{itemize}
\item \textbf{EUBOC}: Use our EUBOC acquisition function with warm-starting. We pre-trained the constraint surrogate model using 200 points randomly sampled from the search space.
\item \textbf{EUBOC w/o warm-starting}: Use our EUBOC acquisition function without warm-starting.
\item \textbf{EUBO}: Use the EUBO acquisition function. This method ignores the constraints.
\item \textbf{EUBO w/ cons.}: Use the EUBO acquisition function; however, if only one of the two candidate points satisfies the constraint, we automatically treat it as the winner. This method represents a naive, post-hoc way of handling constraints.
\item \textbf{Random}: Use uniform random sampling. This method ignores the constraints.
\end{itemize}

\subsubsection{Performance Metrics}
\label{sec:metrics}

The performance metrics for the evaluation are as follows:
\begin{itemize}
\item \emph{Optimality gap}: The difference between the optimal function value and the best-found function value~\cite{Wang2016-gz,Koyama2020-pm}. We set the best-found function value to the worst (smallest) objective function value if only infeasible values are observed (following \cite{Hernandez-Lobato2015-dm,Lam2017-xc}).
\item \emph{Feasible}: The cumulative proportion of sample points that satisfy the constraints up to each iteration (following \cite{Gardner2014-tv}).
\end{itemize}
The number of iterations was set to 50.
We ran each method 50 times with random initializations and recorded the mean and standard deviation of the results of all runs.

\subsubsection{Results}

\autoref{fig:synthetic_results} shows the performance of each method.
Overall, we find that our proposed EUBOC and EUBOC w/o warm-starting methods consistently outperformed the other methods across all settings.
In addition, compared to EUBOC w/o warm-starting, EUBOC demonstrated a significant performance improvement thanks to the pre-training of the constraint surrogate model (see \autoref{sec:appendix:warm_eval} for further analysis).

As shown in \autoref{fig:synthetic_results} (a), for the 2D test function, the EUBOC methods could quickly reduce the \emph{Optimality gap}, whereas both EUBO and EUBO w/ cons.\ reduced the gap only slowly.
EUBO struggled because it focused much on sampling constraint-violating regions.
EUBO w/ cons., while slightly improved by considering the constraint, did not achieve significant gains, probably due to its naive constraint handling.
Furthermore, EUBOC shows a smaller standard deviation, indicating robust convergence regardless of initialization.

As shown in \autoref{fig:synthetic_results} (b), EUBOC successfully explored only within the feasible region throughout the iterations in the 2D setting, thanks to warm-starting and the constraint-aware acquisition function.
EUBOC w/o warm-starting and EUBO w/ cons.\ gradually increased the \emph{Feasible} during iterations.
EUBOC w/o warm-starting, which explicitly learns the constraint, could explore the feasible region more quickly.

As shown in \autoref{fig:synthetic_results} (c) and (d), for the 6D test function, EUBOC effectively focused on feasible regions in almost every iteration, efficiently reducing the \emph{Optimality gap} compared to the other methods, similar to the performance in 2D.
EUBOC w/o warm-starting also demonstrated its ability to learn the constraint during iterations and reduced the gap effectively.

\subsection{Banner Ad Design Application Setting}
\label{sec:poc_setting}

We next simulated human responses to evaluate the effectiveness of our proposed technique in a practical banner ad design context.
Our application system, described in \autoref{sec:poc}, adjusts the colors (or layouts) of target banner ad images while ensuring a minimum required click-through rate (CTR) predicted by a machine learning model.

\subsubsection{Task}
The task was to adjust the color of a banner ad image so that it closely matched a predetermined reference image as possible.
The reference image was created by recoloring an original banner ad, using parameter values selected from the feasible design space (see \autoref{fig:banner_reference} for the reference image).
We set the threshold $\lambda$ (\ie, the minimum acceptable predicted CTR) to the average of CTR values obtained from 1,000 randomly sampled parameters.
The target image was described by six parameters (\ie, a 6D parameter space).
At each iteration, the ``human response'' to any pairwise comparison was synthesized so that the chosen image would be closer to the reference image than its alternative.

\subsubsection{Methods to be Compared}
The compared methods were the same as those used in \autoref{sec:synthetic_setting}, except that for the EUBOC setting, we pre-trained the constraint surrogate model using 1,000 random recoloring parameters for the warm start.

\subsubsection{Performance Metrics}
We used the \emph{Image gap} as a performance metric~\cite{Yamamoto2022-vi}.
It is defined as the average of the element-wise absolute difference between the two images, each of which is represented as a tensor in $[0, 255]^{W\times H\times 3}$ ($W$ and $H$ denote the width and height of the image, respectively).
We regarded the \emph{Image gap} value as 255 if only infeasible values were observed.
We also recorded the \emph{Feasible}, the proportion of the images satisfying the constraints (as described in \autoref{sec:metrics}).

\figsystemresults

\subsubsection{Results}

\autoref{fig:system_results} shows the results.
Our EUBOC efficiently reduced the \emph{Image gap} by focusing its sampling on feasible regions (about 90\% of the time).
Consistent with \autoref{sec:synthetic_setting}, EUBOC outperforms others on this problem, which resembles real-world banner ad design tasks.
However, the EUBOC w/o warm-starting did not reduce the \emph{Image gap} effectively, showing similar performance to other baselines.
This is likely due to the complex shape of the constraint, which can be difficult to learn at the early stages of optimization. Warm-starting therefore proves especially beneficial here (see \autoref{sec:appendix:warm_eval} for further analysis).
Although the constraint was not satisfied at every iteration, possibly due to the non-exhaustive pre-training of the constraint surrogate model, the \emph{Feasible} gradually increased as the constraint surrogate model was also updated during the iterations.
See \autoref{fig:ad_results_tech_eval} for examples showing how banner ad images evolved throughout the optimization process.


\section{Application: A Designer-in-the-Loop Banner Ad Design Framework}

In addition to proposing CPBO as a novel optimization method, we apply it to a real-world banner ad design challenge.
This demonstration not only highlights CPBO's practical value but also contributes to the human-in-the-loop design optimization field, where computational methods integrate with human subjective judgment \cite{Brochu2007-xw,Koyama2022-by}.

\subsection{Motivation and Background}
\label{sec:motivation}

\subsubsection{Human-in-the-Loop Optimization for Graphic Design}

\emph{Human-in-the-loop} optimization allows human evaluators---in our case, designers---to act as the objective function and guide the search process \cite{Koyama2017-yz,Yamamoto2022-vi,Brochu2010-qi}.
Researchers have explored such methods for various design tasks \cite{Koyama2022-ec}, including visual design \cite{Koyama2017-yz,Yamamoto2022-vi,Brochu2010-qi,Koyama2020-pm,Koyama2022-by} and interaction design \cite{Khajah2016-cd,Liao2023-pf,Dudley2019-fq,Kadner2021-us}.
Our work adds to this body of Human-Computer Interaction (HCI) research by incorporating an additional \emph{performance-oriented} design constraint---in our case, CTR---into the designer-in-the-loop framework.

\subsubsection{Banner Ad Design Challenges}

Banner ad design poses a unique challenge: achieving visual appeal (\ie, the designer's visual preference) while also maintaining a sufficiently high CTR.
Traditional approaches may require extensive real-world testing to measure actual CTRs, which is both costly and time-consuming.
Moreover, designers cannot reliably predict how visual changes affect CTR, and preference often does not align with actual CTR.
Consequently, a method that \emph{automatically ensures CTR} while letting designers focus on aesthetics is highly desirable.

\subsubsection{Preliminary Study: Designer Preference vs. CTR}
\label{sec:prelim_study}

To inform and motivate our approach, we conducted a \emph{preliminary study} with professional ad designers.
Notably, the study found \emph{no positive correlation} between a designer's preference and the actual CTR.
This finding reinforces our core hypothesis that an automated system (rather than the designer) should manage CTR constraints, freeing the designer to focus on their creative intent.
Detailed procedures and results of this study are provided in \autoref{sec:appendix:preliminary}.

\subsection{Design Framework}

\subsubsection{Concept}

\figureframework

We propose a designer-in-the-loop framework for banner ad design that integrates both aesthetic preferences and CTR considerations (\autoref{fig:framework}).
By adjusting design parameters (\eg, colors, layouts), the system aims to produce visually appealing ads while maintaining a minimum required CTR.
Designers choose their preferred option from a pair of system-generated candidate ad images, focusing on creative decisions, while the system automatically handles CTR constraints using a CTR prediction model\footnote{Measuring actual CTRs is impractical due to the time and financial costs, so our framework uses a machine-learning-based CTR prediction model trained on real-world data instead.}.
This setup addresses the challenge of simultaneously considering both preference and CTR during the design process.

This designer-in-the-loop optimization approach leverages our CPBO technique.
The system maintains surrogate models for both the preference (objective) and CTR (constraint) functions.
After each step of designer feedback and CTR prediction, these models are updated, and a new design candidate pair is selected to maximize preference under the CTR constraint.
Through repeating this iterative process, the framework enables effective human-AI collaboration between the designer and the optimization module, ultimately producing designs that meet both aesthetic and performance goals.

\subsubsection{System Implementation}
\label{sec:poc}

We implemented a proof-of-concept system that supports banner ad design with two separate modes: a \emph{color} editing mode and a \emph{layout} editing mode.
\autoref{fig:user_interface} shows the user interface of our system in color editing mode.
Our system is based on pairwise comparisons and presents the designer with two banner ad designs in each iteration.
Following the designer's selection, the next two candidate designs are presented based on their preference.
This iterative process allows the designer to explore more preferable designs, with the system ensuring CTR.

We implemented our CPBO technique on BoTorch~\cite{Balandat2020-vg}, a BO library.
We used Gaussian process models in BoTorch as the surrogate models for the objective and constraint functions.
We implemented EUBOC by extending the EUBO implementation in BoTorch.

The CTR prediction model\footnote{Note that the CTR prediction model and the CTR surrogate model are different; the prediction model is a fixed model, while the surrogate model is dynamically updated during iterations and used for CPBO computation.} was built using XGBoost~\cite{Chen2016-kq}, trained on our dataset of real-world banner ad deployment data.
This model predicts a CTR value for each given banner ad image, and those predictions update the constraint surrogate model during the optimization process (and optional warm-starting).
Note that CTR prediction takes approximately 0.3 milliseconds per image.

Refer to \autoref{sec:appendix:cpbo} and \autoref{sec:appendix:poc} for more details.

\subsection{User Study}
\label{sec:study}

\figuserinterface

We conducted a user study using our system (\autoref{sec:poc}).
This experiment was carried out with the approval of the ethic examination of Research Institute of Human Engineering for Quality Life.
The goal was to evaluate how our CPBO-enabled framework impacts on the user experience in banner ad design.
Specifically, we aimed to assess the benefits of having a designer-in-the-loop design system responsible for CTR constraint, allowing designers to focus on their preferences.
To this end, we compared two scenarios:
(1) Ours: the system ensures CTR while the designer focuses on preferences, and
(2) Baseline: the system does not ensure CTR.

\subsubsection{Study Design}
\label{sec:study:method}

We recruited 11 professional ad designers (P1--P11) from an ad agency in Japan (10 females, 1 male, age: $M = 28.3, SD = 4.59$).
They had an average of 5.82 years ($SD = 4.79$) of general design experience and 3.55 years ($SD = 3.01$) of specific experience in banner ad design.

We prepared two realistic banner ad images (12D and 6D parameter spaces, respectively) by asking a non-participant professional designer, using Japanese-language content.
Each participant performed color-editing tasks on both images under the two system conditions.
The image and system condition pairings and their conduct order were randomized.
In each task, the participant selected the more preferred option from the pair of color variations the system presented, repeating this 50 times.
(Note: the average time to provide preference feedback was 4.2 seconds, ranging from 1.0 to 23.1 seconds.)
Participants were told to consider only their preference with our system, and both preference and CTR with the baseline.
To reduce the bias from the trust in the model, we informed participants that the CTR prediction model was validated.

After finishing both tasks, participants completed a 7-point Likert-scale questionnaire (7: ``strongly agree'') and could provide free-form comments.
We also conducted semi-structured interviews\footnote{Interviews were conducted in the participants' native language (Japanese); the quoted remarks here are translated.} regarding the tasks performed and questionnaire answers.

\subsubsection{Questionnaire Results}

\figquestionarrie

\autoref{fig:questionarrie} shows the questionnaire results.
Overall, professional ad designers evaluated our framework positively.
Q1--Q3 focus on the CTR consideration (higher is better), while Q4 measures perceived stress (lower is better).
Affirmative responses (\ie, 5--7 for Q1--Q3; 1--3 for Q4) were 63.6\% (Q1), 81.8\% (Q2), 63.6\% (Q3), and 81.8\% (Q4).
Notably, Q1 and Q2, which address how the system manages CTR, received strong support.

\subsubsection{Interview Results}

We summarize the feedback on the following two points.

\paragraph{How the system's CTR consideration is helpful}
We received various positive comments about the system's CTR consideration.
P5 appreciated how it accounts for CTR during the design process because \emph{``as a designer, I want to know what makes an effective banner from a third-party [objective] perspective.''}
P3 noted that the system's help \emph{``reduced the effort needed to consider CTR,''} suggesting an overall reduction in mental workload.
Others also commented on differences in candidate quality compared to the baseline.
P8 thought that the candidates provided by our system \emph{``had better visibility''} and \emph{``avoided eyestrain''}.
P10 complained that, when trying the baseline, \emph{``I got ridiculously bad ones [design candidates] many times,''} making choices trivial; in contrast, with our system, \emph{``I was quite indecisive''} since both candidates were often good.
These comments suggest that the CTR consideration helped to provide more reasonable and meaningful design candidates.

\paragraph{How the system's CTR consideration changes future ad design}
P9 appreciated the use of actual measured data for training, saying \emph{``It is really powerful for us [ad designers] to have the designs [created with ours] backed up by thousands of data.''}
P9 added that being able to \emph{``explain this [mechanism of CTR consideration] [to the client]''} is powerful, making the tool \emph{``branded''} and become \emph{``a persuasive material when pitching to clients.''}
These remarks highlight not only the practical benefits for designers, but also the potential to improve client communication and trust, thus adding business value.

\subsubsection{Lessons Learned}

In our user study, we evaluated the benefits of our approach, where the system takes responsibility for CTR considerations in the design process, assisting ad designers.
Feedback from the questionnaires and interviews showed that professional ad designers responded positively to the framework.
Participants noted that it could be particularly helpful to those struggling to account for CTR or seeking an objective perspective on their designs.

Participants also indicated that incorporating CTR considerations improved the quality of design candidates in the iterative process compared to when CTR was not considered.
This suggests that including the CTR constraint enhances optimization performance, supporting our technical evaluation of CPBO performance (\autoref{sec:tech_eval}).

Finally, participants highlighted that our approach helps explain designs to clients, as they are based on actual data rather than potentially unreliable intuition.
This strengthens communication and trust between designers and clients, demonstrating the business value of data-driven insights in improving both the design process and client relationships.


\section{Discussions and Future Work}

\subsection{Improving Search Efficiency and Capability}
\label{sec:discussion:efficiency}

Our experiments used search spaces of up to 12 dimensions and 50 iteration steps---enough to observe optimization behavior.
In practice, designers may wish to adjust more design elements, leading to even higher-dimensional search spaces, and also minimize the required iterations.
Thus, improving search efficiency is crucial.

High-dimensional BO is known to be challenging, and various methods have been proposed \cite{Wang2016-gz,Binois2022-qp,Long2024-qu,Hvarfner2024-vk};
future work should explore ways to combine these methods with CPBO.
Another promising avenue is to enable designers to compare more than two search points simultaneously in each step~\cite{Koyama2017-yz,Koyama2020-pm,Nguyen2021-nu}.
Although our EUBOC currently only supports the evaluation of two search points at a time, extending it by incorporating the concept of qEUBO~\cite{Astudillo2023-ex}, an EUBO extension capable of sampling multiple search points simultaneously, would be a valuable research direction.

Another future direction is to handle categorical variables.
While our EUBOC focuses on continuous inputs, we believe it can \emph{theoretically} be extended to accommodate categorical variables---for example, by incorporating kernel adaptations for GPs~\cite{Garrido-Merchan2020-vo}.

\subsection{Toward Practical Design Systems}

The purpose of developing our application system was to investigate the potential benefits of a CPBO-driven design framework.
The next step is to build a more comprehensive design system for professional use.
Feedback from professional designers during our interviews provided multiple suggestions for improving the system.
They expressed a desire to adjust not only colors or layouts, but also other design elements such as text width, font size, and letter kerning.
Some designers wanted to compare more than two design candidates at a time to make a better decision (as discussed in \autoref{sec:discussion:efficiency}).
In addition, some mentioned that, in their experience, the color and layout directions are often predetermined to some extent (\eg, by client requests) before design exploration begins, suggesting the need for a feature that limits the search space to accommodate these prior intentions, thereby enhancing the design process.


\section{Conclusion}

This paper proposed CPBO and a new acquisition function, EUBOC, to enable this.
Our technical evaluation showed that our method efficiently reduces the gap toward optimal solutions by focusing on feasible regions.
As a practical CPBO application, we proposed a designer-in-the-loop framework for designing banner ads that integrates CTR considerations.
The user study demonstrated that our framework effectively reduced the design burden and proved its usefulness as a real-world CPBO application.

\section*{Ethical Statement}

There are no significant ethical concerns regarding our CPBO.
However, the use of CTR models in banner ad design poses potential risks, such as generating overly attention-grabbing designs that may unnecessarily encourage user clicks.
Our framework mitigates these risks by facilitating collaboration between designers and the algorithm, allowing for designs that align with human subjective preferences.
While this reduces the likelihood of harmful outcomes, further efforts to establish safeguards and ethical guidelines would enhance the robustness of such systems in the future.

\section*{Acknowledgements}

We would like to thank Kazuhito Nishimura and his team from Hakuhodo DY ONE Inc. for their generous participation in our user study and for providing valuable insights that greatly contributed to this work.

\bibliographystyle{named}
\bibliography{bibliography/hitl_ijcai2025}

\appendix

\section{CPBO Implementation Details}
\label{sec:appendix:cpbo}

\subsection{Gaussian Proccess Model}

We used GP models in BoTorch~\cite{Balandat2020-vg} as the surrogate models.
More specifically, \texttt{PairwiseGP} and \texttt{SingleTaskGP} were used for objective and constraint functions, respectively.

A GP model is characterized by its prior mean function and its kernel function.
We set these functions for the models to those used by default in BoTorch.
In \texttt{PairwiseGP}, the prior mean function is a non-zero constant prior mean function, and the kernel function is the radial basis function (RBF) kernel with scaling output variable.
In \texttt{SingleTaskGP}, the prior mean function is a non-zero constant prior mean function, and the kernel function is the Mat\'ern $5/2$ kernel with scaling output variable.
The kernel hyperparameters were optimized from the data by the logarithm of marginal likelihood maximization, assuming a gamma prior distribution (standard in BoTorch).

\subsection{Acquisition Function}

EUBOC is based on BoTorch's EUBO implementation, and we extended it to multiply the feasibility of constraints.
Our CPBO samples two points in each step by maximizing the EUBOC.
We used L-BFGS-B~\cite{Byrd1995-kw} for this maximization.
Acquisition functions are often difficult to optimize as they are generally non-convex, so BoTorch makes use of multiple random restarts to improve optimization quality.
For the restarts settings, the number of initial points (\texttt{num\_restarts}) was set to 3, and the number of sampling points to select initial points (\texttt{raw\_samples}) was set to 512.

\section{Technical Evaluation Details}
\label{sec:appendix:tech_eval}

\figwarmstartresults

\subsection{Test Functions}
\label{sec:appendix:synth}

We used two different test functions in our technical evaluation.

The first test function is the 2D test function that comes from Gardner \textit{et al.}~\shortcite{Gardner2014-tv}.
Specifically, the objective function is
\begin{equation}
f_1(x_1,x_2)=-\cos(2x_1)\cos(x_2)-\sin(x_1),
\end{equation}
and the constraint function is
\begin{equation}
c_1(x_1,x_2)=-\cos(x_1)\cos(x_2)+\sin(x_1)\sin(x_2).
\end{equation}
The inequality constraint is set to $c_1(x_1,x_2) \geq 0.5$, and the search space is defined as $x_1,x_2 \in [0.0, 6.0]$.
The optimal function value is $1.88875$, and the smallest function value is $-2.0$.

The second test function is the 6D test function that comes from Letham \textit{et al.}~\shortcite{Letham2019-sm}.
Specifically, the objective function is based on the Hartmann 6 function~\cite{Picheny2013-kf}:
\begin{equation}
f_2(\bm{x})=\sum_{i=1}^{4}\alpha_i\exp\left(-\sum_{j=1}^{6}{A_{ij}(\bm{x}_j-P_{ij})^2}\right),
\end{equation}
where
\begin{align}
\alpha &= [1.0,1.2,3.0,3.2],\\
A &=
\begin{bmatrix}
10 & 3 & 17 & 3.5 & 1.7 & 8 \\
0.05 & 10 & 17 & 0.1 & 8 & 14 \\
3 & 3.5 & 1.7 & 10 & 17 & 8 \\
17 & 8 & 0.05 & 10 & 0.1 & 14 \\
\end{bmatrix},\\
P &= 10^{-4}
\begin{bmatrix}
1312 & 1696 & 5569 & 124 & 8283 & 5886 \\
2329 & 4135 & 8307 & 3736 & 1004 & 9991 \\
2348 & 1451 & 3522 & 2883 & 3047 & 6650 \\
4047 & 8828 & 8732 & 5743 & 1091 & 381 \\
\end{bmatrix}.
\end{align}
The constraint function uses the norm of $\bm{x}$; that is
\begin{equation}
c_2(\bm{x})=-\|\bm{x}\|.
\end{equation}
The inequality constraint is set to $c_2(\bm{x}) \geq -1.0$, and the search space is defined as $\bm{x} \in [0.0, 1.0]^6$.
The optimal function value is $3.32237$, and the smallest function value is $0.0$.

Note that the original problems of these two~\cite{Gardner2014-tv,Letham2019-sm} are defined as constrained \emph{minimization} problems, where the constraint is defined by ``$\le$'' (not ``$\ge$'').
To adapt these problems to our setting, where we treat \emph{maximization} with a constraint defined by ``$\ge$'', we inverted the signs of both their objective and constraint functions.

\subsection{Effect of the Number of Warm-Start Points}
\label{sec:appendix:warm_eval}

To assess the sensitivity of our EUBOC to the number of warm-start points, we conducted an additional experiment using a reduced set of only 50 warm-start points.
As shown in \autoref{fig:warm-start_results}, EUBOC maintained strong performance under this condition, comparable to the setting with more warm-start data.
This result suggests that the constraint surrogate model can be trained effectively even with relatively little data, demonstrating the robustness of our approach with respect to warm-start size.

\subsection{Ad Designs in Technical Evaluation}
\label{sec:appendix:system_eval}

\figreferenceimage

\figtechadresults

\autoref{fig:banner_reference} displays the reference image used in our technical evaluation in the banner ad design application setting (see \autoref{fig:banner} (b) for the original image).
\autoref{fig:ad_results_tech_eval} displays examples of the results in the banner ad design application setting.
The numbers below the images represent the \emph{Image gap}; the left number for the top image and the right number for the bottom image (lower is better).
An image gap of 255 indicates that the corresponding image does not satisfy the constraint.

\section{Banner Ad Design System Details}
\label{sec:appendix:poc}

\subsection{User Interface}

\figuserinterfaceadd

\autoref{fig:user_interface_add} (a) and (b) show the user interface of our system in color editing mode and layout editing mode, respectively.
Our system is based on pairwise comparisons and presents the designer with two banner ad designs.
Following the designer's selection, two new candidate designs are presented based on their preference.
This iterative process allows the designer to explore more preferable designs, with the system ensuring CTR.

\subsection{CTR Prediction Model}
\label{sec:appendix:pctr_model}

Each iteration of CPBO needs to evaluate the CTR values of the design candidates, and we employ a machine learning model to predict the CTRs, instead of actually measuring the CTRs through an ad delivery.
We use XGBoost~\cite{Chen2016-kq} to construct a CTR regression model based on gradient-boosting decision trees~\cite{Friedman2001-jd}.
Our CTR prediction model takes a banner ad image as a 1,000-dimensional input vector extracted by a pre-trained EfficientNetV2~\cite{Tan2021-er} model\footnote{\url{https://github.com/huggingface/pytorch-image-models}}, and then outputs its predicted CTR value.
The model was trained on our in-house dataset, consisting of pairs of banner ad images and corresponding actual measured CTRs.
Each CPBO iteration involves two design candidates, and the two corresponding predicted CTR values are obtained by this model.
These predicted CTR values are then used to update the surrogate model $c$.

\subsection{Color Editing Technique}

We adopt a \emph{palette-based recoloring} approach~\cite{Chang2015-av,Tan2018-kz} to change the colors of banner ad images in a parametric way.
More specifically, we adopt the technique of Tan \textit{et al.}~\shortcite{Tan2018-kz} as it works sufficiently fast for the use in interactive systems like ours.
Given an input image, this technique creates a \emph{color palette} (\ie, a small set of colors that digest the full range of colors in the image~\cite{Chang2015-av}) and then allows the user to easily recolor the entire image by simply altering the palette colors.
In our case, it is the CPBO system, not the user, that alters the palette colors.
Note that this technique can successfully recolor images with complex gradients (such as photographs) while we apply it to typical ad banner images primarily consisting of flat colors (refer to \autoref{fig:banner}).

Each color in the palette has three parameters (RGB), and so the number of adjustable parameters is $3M$, where $M$ is the size of the palette (typically, $M = 4, \ldots, 7$).
All of these are potentially considered as our search parameters.
However, we found that changing the brightest and darkest colors in the extracted palette is unreasonable as it often produces unnatural results for ad images.
Thus, our system keeps these two colors unchanged but allows changes to other colors in the palette.
As a result, the search space of CPBO becomes $3(M-2)$, accounting for the two fixed colors.
Note that the palette-based technique has the option to automatically determine the number of palette colors, but our implementation allows the user to directly specify the palette size for each image.

\subsection{Layout Editing Technique}

We consider a banner ad layout that consists of a background image and $E$ design elements, such as a product image or text labels for product names.
We assume each element is a rectangle with a fixed aspect ratio, \ie, it is fully represented by its center coordinates ($x, y$) and size scale $s$.
Thus, the overall layout is specified by the parameter set ${(x_1, y_1, s_1), (x_2, y_2, s_2), \ldots, (x_E, y_E, s_E)}$.
The CPBO directly treats these $3E$ parameters as search variables, resulting in a $3E$-dimensional search space.

However, this na\"{i}ve search space includes many infeasible designs due to issues such as element misalignment and overlap, making the search process cumbersome and inefficient as it involves evaluating numerous unsuitable designs.
To improve the feasibility of the search space by eliminating infeasible designs, we apply the constrained layout optimization technique proposed by Kikuchi \textit{et al.}~\shortcite{Kikuchi2021-rm} as a postprocessing step.
Before starting the design session, we specify all the fundamental design constraints for the target banner ad design, such as alignment between specific elements, overlap avoidance, and, if applicable, client requests.
Then, each time the CPBO process selects a new layout, we run Kikuchi \textit{et al.}'s constrained optimization method, using the selected layout as the initial solution.
In this optimization, the objective function is the Euclidean distance from the initial layout, which we aim to minimize within the specified constraints.
The result of this optimization satisfies all specified constraints while remaining as close as possible to the initial solution.

\section{Preliminary Study Details}
\label{sec:appendix:preliminary}

\subsection{Goal}

To gain a better understanding of the task of banner ad design and validate our motivation, we surveyed 12 professional ad designers (D1, \ldots, D12).
This study aims to (1) verify whether designers can accurately predict CTR (pCTR) and (2) examine whether there is a correlation between designers' preferences and actual CTR (aCTR).

\subsection{Method}

\begin{table}[t]
    \caption{Data obtained from our preliminary study conducted with professional ad designers.}
    \label{tab:questionnaire_answers}
    \centering
    \small
    \begin{tabular}{ccccc}
    \toprule
     Designer & Banner set & Preference & pCTR & aCTR \\
    \midrule
     D1 & Set1 &
     $\mathrm{A}\!>\!\mathrm{B}\!>\!\mathrm{C}$ &
     $\mathrm{A}\!>\!\mathrm{B}\!>\!\mathrm{C}$ &
     $\mathrm{C}\!>\!\mathrm{B}\!>\!\mathrm{A}$ \\
     D1 & Set2 &
     $\mathrm{B}\!>\!\mathrm{A}\!>\!\mathrm{C}$ &
     $\mathrm{B}\!>\!\mathrm{A}\!>\!\mathrm{C}$ &
     $\mathrm{C}\!>\!\mathrm{B}\!>\!\mathrm{A}$ \\
     D1 & Set3 &
     $\mathrm{B}\!>\!\mathrm{C}\!>\!\mathrm{A}$ &
     $\mathrm{B}\!>\!\mathrm{C}\!>\!\mathrm{A}$ &
     $\mathrm{A}\!>\!\mathrm{C}\!>\!\mathrm{B}$ \\
     D2 & Set1 &
     $\mathrm{C}\!>\!\mathrm{B}\!>\!\mathrm{A}$ &
     $\mathrm{B}\!>\!\mathrm{C}\!>\!\mathrm{A}$ &
     $\mathrm{C}\!>\!\mathrm{B}\!>\!\mathrm{A}$ \\
     D2 & Set2 &
     $\mathrm{A}\!>\!\mathrm{B}\!>\!\mathrm{C}$ &
     $\mathrm{A}\!>\!\mathrm{B}\!>\!\mathrm{C}$ &
     $\mathrm{C}\!>\!\mathrm{B}\!>\!\mathrm{A}$ \\
     D2 & Set3 &
     $\mathrm{B}\!>\!\mathrm{A}\!>\!\mathrm{C}$ &
     $\mathrm{A}\!>\!\mathrm{B}\!>\!\mathrm{C}$ &
     $\mathrm{A}\!>\!\mathrm{C}\!>\!\mathrm{B}$ \\
     D3 & Set1 &
     $\mathrm{A}\!>\!\mathrm{B}\!>\!\mathrm{C}$ &
     $\mathrm{B}\!>\!\mathrm{A}\!>\!\mathrm{C}$ &
     $\mathrm{C}\!>\!\mathrm{B}\!>\!\mathrm{A}$ \\
     D3 & Set2 &
     $\mathrm{A}\!>\!\mathrm{B}\!>\!\mathrm{C}$ &
     $\mathrm{A}\!>\!\mathrm{B}\!>\!\mathrm{C}$ &
     $\mathrm{C}\!>\!\mathrm{B}\!>\!\mathrm{A}$ \\
     D3 & Set3 &
     $\mathrm{C}\!>\!\mathrm{A}\!>\!\mathrm{B}$ &
     $\mathrm{C}\!>\!\mathrm{B}\!>\!\mathrm{A}$ &
     $\mathrm{A}\!>\!\mathrm{C}\!>\!\mathrm{B}$ \\
     D4 & Set1 &
     $\mathrm{C}\!>\!\mathrm{A}\!>\!\mathrm{B}$ &
     $\mathrm{A}\!>\!\mathrm{C}\!>\!\mathrm{B}$ &
     $\mathrm{C}\!>\!\mathrm{B}\!>\!\mathrm{A}$ \\
     D4 & Set2 &
     $\mathrm{C}\!>\!\mathrm{A}\!>\!\mathrm{B}$ &
     $\mathrm{A}\!>\!\mathrm{C}\!>\!\mathrm{B}$ &
     $\mathrm{C}\!>\!\mathrm{B}\!>\!\mathrm{A}$ \\
     D4 & Set3 &
     $\mathrm{A}\!>\!\mathrm{B}\!>\!\mathrm{C}$ &
     $\mathrm{B}\!>\!\mathrm{A}\!>\!\mathrm{C}$ &
     $\mathrm{A}\!>\!\mathrm{C}\!>\!\mathrm{B}$ \\
     D5 & Set1 &
     $\mathrm{B}\!>\!\mathrm{C}\!>\!\mathrm{A}$ &
     $\mathrm{B}\!>\!\mathrm{A}\!>\!\mathrm{C}$ &
     $\mathrm{C}\!>\!\mathrm{B}\!>\!\mathrm{A}$ \\
     D5 & Set2 &
     $\mathrm{A}\!>\!\mathrm{B}\!>\!\mathrm{C}$ &
     $\mathrm{B}\!>\!\mathrm{C}\!>\!\mathrm{A}$ &
     $\mathrm{C}\!>\!\mathrm{B}\!>\!\mathrm{A}$ \\
     D5 & Set3 &
     $\mathrm{C}\!>\!\mathrm{A}\!>\!\mathrm{B}$ &
     $\mathrm{A}\!>\!\mathrm{C}\!>\!\mathrm{B}$ &
     $\mathrm{A}\!>\!\mathrm{C}\!>\!\mathrm{B}$ \\
     D6 & Set1 &
     $\mathrm{A}\!>\!\mathrm{C}\!>\!\mathrm{B}$ &
     $\mathrm{A}\!>\!\mathrm{C}\!>\!\mathrm{B}$ &
     $\mathrm{C}\!>\!\mathrm{B}\!>\!\mathrm{A}$ \\
     D6 & Set2 &
     $\mathrm{A}\!>\!\mathrm{C}\!>\!\mathrm{B}$ &
     $\mathrm{B}\!>\!\mathrm{A}\!>\!\mathrm{C}$ &
     $\mathrm{C}\!>\!\mathrm{B}\!>\!\mathrm{A}$ \\
     D6 & Set3 &
     $\mathrm{A}\!>\!\mathrm{C}\!>\!\mathrm{B}$ &
     $\mathrm{A}\!>\!\mathrm{C}\!>\!\mathrm{B}$ &
     $\mathrm{A}\!>\!\mathrm{C}\!>\!\mathrm{B}$ \\
     D7 & Set1 &
     $\mathrm{C}\!>\!\mathrm{A}\!>\!\mathrm{B}$ &
     $\mathrm{C}\!>\!\mathrm{A}\!>\!\mathrm{B}$ &
     $\mathrm{C}\!>\!\mathrm{B}\!>\!\mathrm{A}$ \\
     D7 & Set2 &
     $\mathrm{A}\!>\!\mathrm{C}\!>\!\mathrm{B}$ &
     $\mathrm{B}\!>\!\mathrm{C}\!>\!\mathrm{A}$ &
     $\mathrm{C}\!>\!\mathrm{B}\!>\!\mathrm{A}$ \\
     D7 & Set3 &
     $\mathrm{C}\!>\!\mathrm{B}\!>\!\mathrm{A}$ &
     $\mathrm{C}\!>\!\mathrm{B}\!>\!\mathrm{A}$ &
     $\mathrm{A}\!>\!\mathrm{C}\!>\!\mathrm{B}$ \\
     D8 & Set1 &
     $\mathrm{A}\!>\!\mathrm{C}\!>\!\mathrm{B}$ &
     $\mathrm{A}\!>\!\mathrm{B}\!>\!\mathrm{C}$ &
     $\mathrm{C}\!>\!\mathrm{B}\!>\!\mathrm{A}$ \\
     D8 & Set2 &
     $\mathrm{A}\!>\!\mathrm{B}\!>\!\mathrm{C}$ &
     $\mathrm{C}\!>\!\mathrm{A}\!>\!\mathrm{B}$ &
     $\mathrm{C}\!>\!\mathrm{B}\!>\!\mathrm{A}$ \\
     D8 & Set3 &
     $\mathrm{B}\!>\!\mathrm{A}\!>\!\mathrm{C}$ &
     $\mathrm{C}\!>\!\mathrm{A}\!>\!\mathrm{B}$ &
     $\mathrm{A}\!>\!\mathrm{C}\!>\!\mathrm{B}$ \\
     D9 & Set1 &
     $\mathrm{A}\!>\!\mathrm{B}\!>\!\mathrm{C}$ &
     $\mathrm{A}\!>\!\mathrm{C}\!>\!\mathrm{B}$ &
     $\mathrm{C}\!>\!\mathrm{B}\!>\!\mathrm{A}$ \\
     D9 & Set2 &
     $\mathrm{B}\!>\!\mathrm{C}\!>\!\mathrm{A}$ &
     $\mathrm{C}\!>\!\mathrm{A}\!>\!\mathrm{B}$ &
     $\mathrm{C}\!>\!\mathrm{B}\!>\!\mathrm{A}$ \\
     D9 & Set3 &
     $\mathrm{B}\!>\!\mathrm{C}\!>\!\mathrm{A}$ &
     $\mathrm{B}\!>\!\mathrm{A}\!>\!\mathrm{C}$ &
     $\mathrm{A}\!>\!\mathrm{C}\!>\!\mathrm{B}$ \\
     D10 & Set1 &
     $\mathrm{C}\!>\!\mathrm{B}\!>\!\mathrm{A}$ &
     $\mathrm{C}\!>\!\mathrm{A}\!>\!\mathrm{B}$ &
     $\mathrm{C}\!>\!\mathrm{B}\!>\!\mathrm{A}$ \\
     D10 & Set2 &
     $\mathrm{A}\!>\!\mathrm{C}\!>\!\mathrm{B}$ &
     $\mathrm{C}\!>\!\mathrm{A}\!>\!\mathrm{B}$ &
     $\mathrm{C}\!>\!\mathrm{B}\!>\!\mathrm{A}$ \\
     D10 & Set3 &
     $\mathrm{C}\!>\!\mathrm{A}\!>\!\mathrm{B}$ &
     $\mathrm{A}\!>\!\mathrm{C}\!>\!\mathrm{B}$ &
     $\mathrm{A}\!>\!\mathrm{C}\!>\!\mathrm{B}$ \\
     D11 & Set1 &
     $\mathrm{A}\!>\!\mathrm{C}\!>\!\mathrm{B}$ &
     $\mathrm{A}\!>\!\mathrm{C}\!>\!\mathrm{B}$ &
     $\mathrm{C}\!>\!\mathrm{B}\!>\!\mathrm{A}$ \\
     D11 & Set2 &
     $\mathrm{C}\!>\!\mathrm{A}\!>\!\mathrm{B}$ &
     $\mathrm{A}\!>\!\mathrm{C}\!>\!\mathrm{B}$ &
     $\mathrm{C}\!>\!\mathrm{B}\!>\!\mathrm{A}$ \\
     D11 & Set3 &
     $\mathrm{C}\!>\!\mathrm{B}\!>\!\mathrm{A}$ &
     $\mathrm{C}\!>\!\mathrm{B}\!>\!\mathrm{A}$ &
     $\mathrm{A}\!>\!\mathrm{C}\!>\!\mathrm{B}$ \\
     D12 & Set1 &
     $\mathrm{A}\!>\!\mathrm{B}\!>\!\mathrm{C}$ &
     $\mathrm{A}\!>\!\mathrm{B}\!>\!\mathrm{C}$ &
     $\mathrm{C}\!>\!\mathrm{B}\!>\!\mathrm{A}$ \\
     D12 & Set2 &
     $\mathrm{A}\!>\!\mathrm{B}\!>\!\mathrm{C}$ &
     $\mathrm{B}\!>\!\mathrm{A}\!>\!\mathrm{C}$ &
     $\mathrm{C}\!>\!\mathrm{B}\!>\!\mathrm{A}$ \\
     D12 & Set3 &
     $\mathrm{B}\!>\!\mathrm{A}\!>\!\mathrm{C}$ &
     $\mathrm{B}\!>\!\mathrm{A}\!>\!\mathrm{C}$ &
     $\mathrm{A}\!>\!\mathrm{C}\!>\!\mathrm{B}$ \\
    \bottomrule
    \end{tabular}
\end{table}

\begin{table}[t]
    \caption{Rankings, $d_i$, and $d_i^2$ of pCTR and aCTR in D1 and Set1.}
    \label{tab:rs_sample}
    \centering
    \begin{tabular}{ccccc}
    \toprule
     Banner & pCTR & aCTR & $d_i$ & $d_i^2$ \\
    \midrule
     $\mathrm{A}$ & $1$ & $3$ & $-2$ & $4$ \\
     $\mathrm{B}$ & $2$ & $2$ & $0$  & $0$ \\
     $\mathrm{C}$ & $3$ & $1$ & $2$  & $4$ \\
    \bottomrule
    \end{tabular}
\end{table}

We prepared three banner ad sets (Set1, Set2, and Set3).
Each set consisted of three banner ads ($\mathrm{A}$, $\mathrm{B}$, and $\mathrm{C}$) with minor variations in design elements, such as color and layout, while maintaining consistency among other conditions (\eg, campaign, target audience, ad image size).
We asked the designers to provide ranking orders of banner ads (\ie, $\mathrm{A}$, $\mathrm{B}$, and $\mathrm{C}$) for every set, based on (1) their preferences and (2) their pCTRs.
Furthermore, the aCTRs for these banner ads were obtained when actually delivered to the internet prior to the study.
As a result, we obtained $36 \: (= 12 \times 3)$ sets of responses on preference, pCTR, and aCTR rankings from the 12 designers across the three banner ad sets.
\autoref{tab:questionnaire_answers} shows the data obtained from our preliminary study.

We calculated Spearman's rank correlation coefficients ($r_s$) for the combinations of pCTR and aCTR rankings and the combinations of the preference and aCTR rankings, respectively.
The following equation expresses Spearman's rank correlation coefficient $r_s$:
\begin{equation}
    r_s = 1 - \frac{6\sum_{i=1}^n d_i^2}{n(n^2-1)},
    \label{eq:rs}
\end{equation}
where $n$ is the number of rankings ($= 3$ in our case), and $d_i$ is the difference between the two rankings of each observation.
\autoref{tab:rs_sample} shows the rankings, $d_i$, and $d_i^2$ of pCTR and aCTR in D1 and Set1.
Applying to \autoref{eq:rs} in this case, we obtain $r_s = -1.0$.
Without loss of generality, if we fix the rankings of pCTR for A, B, and C as 1, 2, and 3, respectively, we have 6 (\ie, $3!$) rankings patterns of a CTR.
We define $S = \sum_{i=1}^3 d_i^2$, then the ranking of aCTR for A, B, and C, $S$, and $r_s$ can be classified into the following four patterns:
\begin{itemize}
    \item When the ranking is 1, 2, 3, then $S=0$ and $r_s=1.0$.
    \item When the ranking is either 2, 1, 3 or 1, 3, 2, then $S=2$ and $r_s=0.5$.
    \item When the ranking is either 2, 3, 1 or 3, 1, 2, then $S=6$ and $r_s=-0.5$.
    \item When the ranking is 3, 2, 1, then $S=8$ and $r_s=-1.0$.
\end{itemize}
Thus, $r_s$ calculated from the data can take one of the following four values: $-1.0$, $-0.5$, $0.5$, or $1.0$.
Values closer to $1.0$ indicate a higher concordance in rankings, with $1.0$ representing perfect rank order agreement and $-1.0$ representing perfect reverse order.

\subsection{Results}

\figquestionnairers

\autoref{fig:questionnaire_rs} shows histograms of the 36 data for Spearman's rank correlation coefficients.
As shown in \autoref{fig:questionnaire_rs} (a), there was no clear correlation between pCTR and aCTR.
However, 63.9\% (23/36) of the total data showed a negative correlation, and surprisingly, only 8.3\% (3/36) correctly predicted the ranking of aCTR.
Similarly, no clear correlation was found between preference and aCTR in \autoref{fig:questionnaire_rs} (b), with 63.9\% (23/36) of the total data showing a negative correlation.

\subsection{Findings}

These results indicate that professional ad designers often struggle to accurately predict how design changes will affect CTR.
Furthermore, the preferences of designers do not positively correlate with the actual CTR, and focusing only on the preferences may negatively impact the actual CTR.
These findings validate our motivation to solve this problem by investigating a design framework in which the system ensures CTR instead of designers.
Our proposed framework aims to address this issue by integrating designers' preferences and the system-predicted CTR into the design optimization process.

\section{User Study Details}
\label{sec:appendix:study}

\subsection{Ad Designs in User Study}
\label{sec:appendix:study:design}

\figbanner

We prepared two types of realistic banner ad images\footnote{These images (originally in Japanese) read (a) ``New Arrival, High Cost-Performance Smartphone with Large Display / MobilePHONE / Product Price 11,980 JPY / Buy Now at Our Online Shop.'', (b) ``Just Plug into the Outlet! / High-Speed Wi-Fi / No Installation Needed / No Data Caps / No Cancellation Fees / Now with Up to 2 Months Free Campaign! / Click Here for Details.''} as displayed in \autoref{fig:banner}.
These ad images were created specifically for this study by a professional ad designer who was not a participant in the study, using Japanese-language content.
The parameter dimension, $3(M-2)$, was 12 ($M = 6$) for \autoref{fig:banner} (a) and 6 ($M = 4$) for \autoref{fig:banner} (b), respectively.

\autoref{fig:ad_results_user_eval} shows an example of ad designs presented and the designer's choice in our user evaluation conducted by P8 and P1.

\figuseradresults

\subsection{Follow-Up User Study}
\label{sec:appendix:follow_up}

The main user evaluation focused only on color as the design parameter, leaving open the question of whether our proposed framework (\ie, CPBO-based, designer-in-the-loop optimization considering CTR) functions effectively for other design parameters.
Additionally, it remained to be seen how designers would respond to the framework when applied to tasks beyond color.
To provide preliminary insights into the framework's broader applicability, we conducted a small, informal follow-up user study centered on layout editing.
This experiment was carried out with the approval of the ethic examination of Research Institute of Human Engineering for Quality Life.
We recruited 6 participants from the original study (P2, P4, P6, P8, P10, and P11).
We created elements of a banner ad for our follow-up user study, as shown in \autoref{fig:ad_follow_up}, based on \autoref{fig:banner} (a).
Using our banner ad design system, the participants performed a layout editing task on the elements of the banner image.
Then, we held semi-structured interviews.

\autoref{fig:ad_results_follow_up} shows an example of ad designs presented and the designer’s choice in our follow-up user study conducted by P4 and P6.

Feedback from the participants suggests that our approach could be particularly effective for layout design tasks.
P10 noted that in typical software, \emph{``Changing colors can be done with a few clicks, but changing the layout requires many more operations, making it more labor-intensive.''}
In contrast, our system automatically generates layout candidates, freeing designers from tedious layout-related operations and facilitating layout exploration.
This view was shared by other participants.
Also, P6 and P11 both emphasized the impact of layout on CTR and appreciated the layout editing capability of our system, with P11 stating, \emph{``layout has a greater effect on CTR as it guides the viewer's eye,''} and P6 added, \emph{``when designing, I usually start with the layout.''}
P2 noted that in long-term client projects, layouts often become stagnant, and expressed a desire for the system to provide \emph{``fresh layout suggestions''} to inspire new ideas, as our system does.

Also, as in the main study, we received positive feedback on the usefulness of the system's CTR consideration.
P6 mentioned that they often need to \emph{``subjectively judge the CTR,''} and found it \emph{``helpful to be able to design with CTR automatically considered.''}
P4 commented on the director's role in CTR consideration, stating, \emph{``It would be useful if designers could account for CTR themselves [using tools],''} as it would reduce the need for back-and-forth with the director.

Overall, the results confirmed that our framework, which had already been validated for color editing, also works effectively for layout editing and was well-received by participants.
This suggests its potential applicability to an even broader range of design parameters, such as font and letter spacing.
One remaining question is whether we should allow for the simultaneous editing of multiple design parameter types;
currently, we separate color and layout editing modes (\ie, users edit either color or layout at a time), but several participants expressed a desire to explore both simultaneously.
We believe there is a trade-off between complexity and flexibility, and exploring this balance will be an important direction toward more practical systems.

\figfollowupbanner

\figfollowupresults

\subsection{Expanding Evaluation in Future Studies}

The purpose of the user evaluation was to demonstrate the concept of our design framework, focusing on the user experience.
As a result, we successfully validated the potential benefits of our approach, encouraging further research in this direction.

Future research should investigate when and how the designer-in-the-loop approach is most appropriate in terms of both designer experience and design performance.
Recently, Chan \textit{et al.}~\shortcite{Chan2022-bp} analyzed the positive and negative aspects of designer-in-the-loop optimization;
similar discussions would be valuable in the presence of complex design constraints, such as CTR in our case.
Additionally, it would be interesting to develop a CPBO-based design system that does not require designers to remain in the optimization loop~\cite{Koyama2022-ec} and compare it with ours in terms of designer experience.

Validating the CTRs of banner ads designed with our framework in an actual online delivery environment would also be beneficial.
However, such experiments are not easy due to monetary, timing, client cooperation, and ethical considerations.
Importantly, this aspect was not our focus because the predicted CTR is assumed to be accurate with the prediction model we used, and evaluating its prediction accuracy is beyond our scope for this study.
It is worth noting that CTR prediction is a highly active research topic, and its advancement is complementary to our work;
our framework is compatible with future CTR prediction models, and the insights gained from the study will remain valid even as new models are developed.

A few designers expressed that they did not fully trust the CTR prediction model (or ``AI'' as they referred to it), although we instructed them that its accuracy had been validated in prior experiments.
Since the trust in CTR prediction models can influence the user experience, evaluating how trust impacts user experience would be an important direction for future work.

\section{Related Work on Graphic Design Support}

Various works have focused on the generation and editing of graphic designs in specific elements such as color and layout.
For color, research has been conducted to aid designers in selecting appropriate color palettes and recoloring~\cite{Shi2023-ci,Hegemann2023-km}, adjusting colors in graphic design photos using a learning-based approach~\cite{Zhao2021-mr}, recoloring photos with a language-based approach~\cite{Wang2023-ol}, and improving the understanding of color editing~\cite{Beaudouin-Lafon2023-yd}.
For layout generation, researchers have employed generative models that consider content and design constraints~\cite{Arroyo2021-by,Zheng2019-hp,Kikuchi2021-rm,Inoue2023-jv,Guo2021-gg}.
Unlike these focused works, our proposed framework is not limited to specific design elements as long as the design is parameterized;
we implemented both color and layout editing capabilities in our proof-of-concept system to demonstrate its generality.

Some works have supported design production by providing design feedback through crowdsourcing~\cite{Xu2014-hw,Luther2015-wb} and saliency analysis~\cite{Bylinskii2017-od}, similar to our work on providing support based on objective design information.
However, our approach differs in that it performs more direct optimization using subjective information (\ie, preference) and objective information (\ie, CTR).
Furthermore, in our framework, since the system is responsible for considering the objective criterion, designers are freed from this task and can focus solely on subjective judgment.

\end{document}